\documentclass[sn-mathphys-num]{sn-jnl}

\usepackage{graphicx}%
\usepackage{multirow}%
\usepackage{amsmath,amssymb,amsfonts}%
\usepackage{amsthm}%
\usepackage{mathrsfs}%
\usepackage[title]{appendix}%
\usepackage{xcolor}%
\usepackage{textcomp}%
\usepackage{manyfoot}%
\usepackage{booktabs}%
\usepackage{algorithm}%
\usepackage{algorithmicx}%
\usepackage{algpseudocode}%
\usepackage{listings}%
\usepackage{xfrac}
\usepackage{acronym}

\usepackage{mathtools}
\usepackage{relsize}
\usepackage{orcidlink}

\DeclareMathOperator*{\argmax}{arg\,max}

\newacro{LEO}{Low Earth Orbit}
\newacro{GEO}{Geostationary Orbit}
\newacro{GTO}{Geostationary Transfer Orbit}
\newacro{SDA}{Space Domain Awareness}
\newacro{AR}{Admissible Region}
\newacro{SLR}{Satellite Laser Ranging}
\newacro{RFS}{Random Finite Set}
\newacro{SSA}{Space Situational Awareness}
\newacro{RMS}{Root Mean Square}
\newacro{PHD}{Probability Hypothesis Density}
\newacro{CPHD}{Cardinalized Probability Hypothesis Density}
\newacro{GMM}{Gaussian Mixture Model}
\newacro{PDF}{Probability Density Function}
\newacro{FOR}{Field of Regard}
\newacro{FOV}{Field of View}
\newacro{PMF}{Probability Mass Function}
\newacro{PIMS}{Perfect Ideal Measurement Set}
\newacro{cell-MB}{cell multi-Bernoulli}
\newacro{POMDP}{Partially-Observed Markov Decision Process}
\newacro{MDP}{Markov Decision Process}
\newacro{KLD}{Kullback-Leibler Divergence}
\newacro{SDT}{Search-Detect-Track}
\newacro{DIS}{Domain Informed Scan}
\newacro{SST}{Space Surveillance Telescope}
\newacro{OD}{Orbit Determination}

\theoremstyle{thmstyleone}%
\theoremstyle{thmstyletwo}%

\theoremstyle{thmstylethree}%

\begin{document}

\title[Information-Driven Search and Track of Novel Space Objects]{Information-Driven Search and Track of Novel Space Objects}


\author*[1]{\fnm{Trevor N.} \sur{Wolf}}\email{twolf@utexas.edu}

\author[1]{\fnm{Brandon A.} \sur{Jones}}\email{brandon.jones@utexas.edu}


\affil*[1]{\orgdiv{Department of Aerospace Engineering and Engineering Mechanics}, \orgname{The University of Texas at Austin}, \orgaddress{\city{Austin}, \postcode{78723}, \state{Texas}, \country{United States of America}}}




\abstract{Space surveillance depends on efficiently directing sensor resources to maintain custody of known catalog objects. However, it remains unclear how to best utilize these resources to rapidly search for and track newly detected space objects. Provided a novel measurement, a search set can be instantiated through admissible region constraints to inform follow-up observations. In lacking well-constrained bounds, this set rapidly spreads in the along-track direction, growing much larger than a follow-up sensor's finite field of view. Moreover, the number of novel objects may be uncertain, and follow-up observations are most commonly corrupted by false positives from known catalog objects and missed detections. In this work, we address these challenges through the introduction of a joint sensor control and multi-target tracking approach. The search set associated to a novel measurement is represented by a Cardinalized Probability Hypothesis Density (CPHD), which jointly tracks the state uncertainty associated to a set of objects and a probability mass function for the true target number. In follow-up sensor scans, the information contained in an empty measurement set, and returns from both novel objects and known catalog objects is succinctly captured through this paradigm. To maximize the utility of a follow-up sensor, we introduce an information-driven sensor control approach for steering the instrument. Our methods are tested on two relevant test cases and we provide a comparative analysis with current naive tasking strategies.}




\maketitle


%
\section{Introduction}\label{introduction}

Rapidly estimating the orbit of newly detected space objects is an open challenge in the \ac{SDA} community. Initial measurements are most commonly provided by optical telescopes. Forming an accurate \ac{OD} solution requires either sufficient geometric diversity between multiple optical tracks or a combination of sensing modalities. Physics-informed approaches such as \ac{AR} constraints allow for generating a representation of dynamically feasible states from single tracks \cite{Milani2004, Fujimoto2013}. Initial \ac{OD} can then be carried out, for instance, by parameterizing the \ac{AR} by a \ac{GMM} and propagating this \ac{PDF} to future observation times at which follow-up measurements resolve the too-short-arc ambiguity \cite{DeMars2013}. Practical situations, however, often lack domain knowledge to place well-constrained bounds in semimajor axis and eccentricity, and as a result, the search space produced with \ac{AR} constraints rapidly spreads in the along-track direction. In this situation, its projection onto the \ac{FOR} can be much larger than a sensor's \ac{FOV}, so multiple sensor scans are likely required to acquire a positive measurement. Depending on the time-out between the initial target detection and follow-up observations, searching this space with a finite FOV may be too cumbersome with bottom-up heuristic sensor tasking strategies \cite{Mahler2014}. Moreover, the projected search set may overlap with known catalog objects, so associating a measurement with the intended target can be ambiguous without rigorous data association procedures. 


In this work, we introduce a closed-loop filtering and control scheme for directing a sensor with a finite \ac{FOV} over a search domain with the intention of rapidly acquiring newly detected space objects. Relating to this, our work makes several unique contributions. First, using tools developed under the \ac{RFS} multi-target tracking paradigm \cite{Mahler2014}, we address the space object acquisition problem while \emph{simultaneously} accounting for 1) multiple targets, 2) measurements corrupted by false-positives and missed detections, and 3) negative information contained in empty measurement scans. Secondly, for directing the steerable sensor, an information-based objective is developed, amenable for our acquisition scenario. This information functional does not rely on the \ac{PIMS} approximation commonly used in the \ac{SDA} literature \cite{Mahler2003}. Finally, a new \ac{GMM} splitting approach is developed, which accurately accounts for the discontinuity in the detection probability between portions inside and outside the \ac{FOV} of the follow-up sensor. This follow-up acquisition approach, while distinct from the \ac{OD} problem, provides a constrained multi-target state estimate which can be used in tertiary tracking algorithms.

Previous work has investigated the acquisition problem in context to acquiring a single target with ideal measurements \cite{Murphy2017}. In \cite{Fedeler2022}, the authors extend these results by incorporating negative information from empty measurements using a conventional single target Bayesian filter. In this work, a set of unknown targets is represented by the \ac{CPHD}. This jointly tracks a posterior estimate of the spatial intensity of targets, parameterized by a \ac{GMM}, in a search domain and a cardinality \ac{PMF} of the true number of targets. Therefore, our approach is adaptable for a multi-target acquisition case. Additionally, we formulate a clutter model based on known catalog objects to account for potential false-positive returns as the domain in searched within the CPHD recursion. This formally updates the intensity and cardinality based on the set of measurements contained in the sensor's finite \ac{FOV}. To that end, at each sensor scan, we include negative information contained in an empty measurement set, varying probabilities of detection over the domain, the potential for clutter returns from known catalog objects, and the information contained in true positive measurements arising from the intended target. 

As the search set is scanned, Gaussian mixands composing the \ac{CPHD} intensity that overlap with the \ac{FOV} bounds must be split to properly account for the discontinuity in the detection probability. To do this, we introduce a Gaussian splitting procedure that splits along optimal principal axes. This approach is similar to that introduced in \cite{LeGrand2022}, however, it is amenable when there exists a nonlinear map between the state and measurement space. Optimal steering of follow-up sensor scans is based on maximizing expected R\'enyi-divergence between the step-wise prior and posterior CPHD \cite{Ristic2011}. This constitutes an information-theoretic approach for sensor management, which have been extensively used in the tracking literature (e.g., \cite{Cai2009, Beard2015, LeGrand2021}). Typically, information functionals do not permit analytic solutions for arbitrary \ac{PDF}s, so approximations are employed. This work introduces a non-parametric approximation of the R\'enyi-divergence based on a weighted $k$-nearest neighbors particle distances that significantly decreases the computational complexity of this objective evaluation in closed-loop sensor control. To demonstrate our methods, we apply them to several realistic scenarios relevant to the \ac{SDA} community and provide comparative results with a typical scanning strategy. 



%
\section{Methodologies}

\subsection{Defining the Search Domain}\label{sec:AR_method}

The \ac{AR} provides a set of dynamically feasible states associated to a single too-short-arc optical measurement track \cite{Milani2004}. This work uses the intersection of \ac{AR} constraints to initialize a search set for propagation and hand-off to follow-up sensor assets. Here, we summarize the \ac{AR} method following that presented in \cite{Worthy2015} and \cite{Gehly2018}. We provide a general mathematical description with applicability to an optical observer assumed. 

A vector-valued function maps a single-target state vector, $\boldsymbol{x} \in \mathbb{R}^{n_x}$, to a measurement vector, $\boldsymbol{z} \in \mathbb{R}^{n_z}$, through
\begin{equation}
    \boldsymbol{z} = \boldsymbol{h}(\boldsymbol{x}; \boldsymbol{p}, t)\mbox{,}\label{eqn:single_targ_measurement}
\end{equation}
where $\boldsymbol{p} \in \mathbb{R}^l$ is a parameter vector (e.g., the observer's state) and $t$ is time. In general, $\boldsymbol{h}(\cdot)$ is rank deficient, so multiple state vectors map to the measurement, $\boldsymbol{z}$. In such an instance, we can find a representation of $\boldsymbol{x}$ partitioned into components determined, $\boldsymbol{x}_d \in \mathbb{R}^{n_d}$, and undetermined, $\boldsymbol{x}_{ud} \in \mathbb{R}^{n_{ud}}$, by the measurement so that
\begin{equation}
    \boldsymbol{x}^\top = [ \boldsymbol{x}_d^\top, \boldsymbol{x}_{ud}^\top ]\mbox{.}
\end{equation}
For example, in the case of an optical measurement, the state vector can be represented in polar coordinates so that the angular bearing and bearing rates are determined, whereas the range and range rate are undetermined. Similarly, we can partition the state argument in Eq. \ref{eqn:single_targ_measurement} so that
\begin{equation}
    \boldsymbol{z} = \boldsymbol{h}(\boldsymbol{x}_d, \boldsymbol{x}_{ud}; \boldsymbol{p}, t)\mbox{.}
\end{equation}
This allows for an inverse map between the measurement and the determinable state components,
\begin{equation}
    \boldsymbol{x}_d = \boldsymbol{h}^{-1}(\boldsymbol{z}; \boldsymbol{p}, t)\mbox{.} \label{eqn:inverse_measurement_function}
\end{equation}
The undetermined state components can be bounded via a set of constraint functions relating to the dynamical characteristics of the observed space object. Let the $i^{\text{th}}$ dynamics constraint function be defined as
\begin{equation}
    d_i(\boldsymbol{x}_d, \boldsymbol{x}_{ud}; \boldsymbol{p}, t) \leq 0\mbox{.}
\end{equation}
Then, substituting Eq. \ref{eqn:inverse_measurement_function},
\begin{equation}
    d_i(\boldsymbol{h}^{-1}(\boldsymbol{z}; \boldsymbol{p}, t), \boldsymbol{x}_{ud}; \boldsymbol{p}, t) \leq 0\mbox{.}
\end{equation}
The set that satisfies the inequality generates the \ac{AR} for the $i^{\text{th}}$ constraint, which can be expressed as
\begin{equation}
    \mathcal{A}_i \coloneqq \{\boldsymbol{x}_{ud} \in \mathbb{R}^{n_{ud}} | d_i(\boldsymbol{h}^{-1}(\boldsymbol{z}; \boldsymbol{p}, t), \boldsymbol{x}_{ud}; \boldsymbol{p}, t) \leq 0 \}\mbox{.} 
\end{equation}
The full \ac{AR} is then simply the intersection of the constrained sets, given by
\begin{equation}
    \mathcal{A} = \mathcal{A}_1 \cap \cdots \cap \mathcal{A}_i \cap \cdots \cap \mathcal{A}_{N_c}\mbox{,}
\end{equation}
where $N_c$ is the number of constraints.

The constraint functions $\{d_i(\cdot)\}_{i = 1}^{N_c}$ are most commonly found by assuming realistic bounds for the eccentricity and specific energy of an elliptical Keplerian orbit. In the case of an optical observer, placing upper and lower bounds on these two quantities produces an admissible region that is closed and bounded. In lacking well-determined bounds, however, the resulting set may contain $\boldsymbol{x}_{ud}$ components that produce orbits intersecting the Earth's surface. To mitigate these unrealistic components, we impose a lower-bound for the periapsis radius, which, depending on the aforementioned eccentricity and energy bounds, may be active. 

In practice, we generate the admissible region through a sufficiently dense, uniform grid of test points placed over the undetermined state components. The constraint functions are sequentially evaluated at each test point, and points not satisfying one or more constraints are discarded. The admissible region is then approximated by a homoscedastic \ac{GMM} by placing mixture components at each admissible test point. The test point density and variance are adapted so that the resulting distribution well-approximates a uniform density over the final \ac{AR}. As noted, for an optical observer, a polar coordinate representation is used for generating the \ac{AR}, however, the resulting \ac{GMM} is then mapped to inertial Cartesian coordinates through the unscented transform for propagation and sensor hand-off \cite{Wan2000}. The propagated \ac{GMM}, when projected into the \ac{FOR}, serves as the search space for follow-up sensor scans. 

\begin{figure}[h]
\centering
\includegraphics[width=1.0\textwidth]{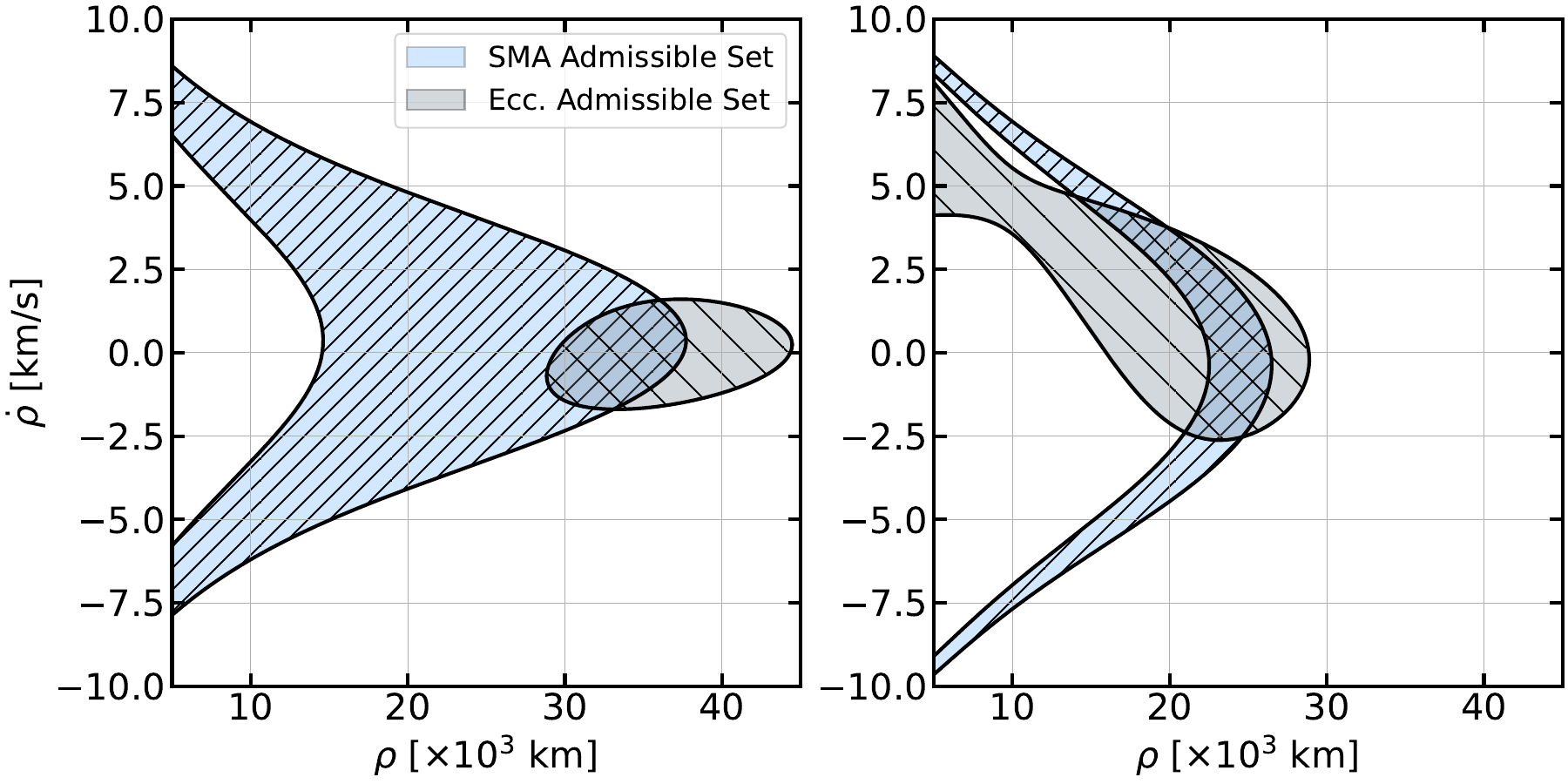}
\caption{\ac{AR} semimajor axis and eccentricity constraints for the \ac{GEO} (left) and \ac{GTO} (right) test cases described in Section \ref{sec:results}. The intersection of the constraint sets is the feasible ranges and range rates associated to the measurement.}\label{fig:AR_figure}
\end{figure}

\begin{figure}[h]
\centering
\includegraphics[width=1.0\textwidth]{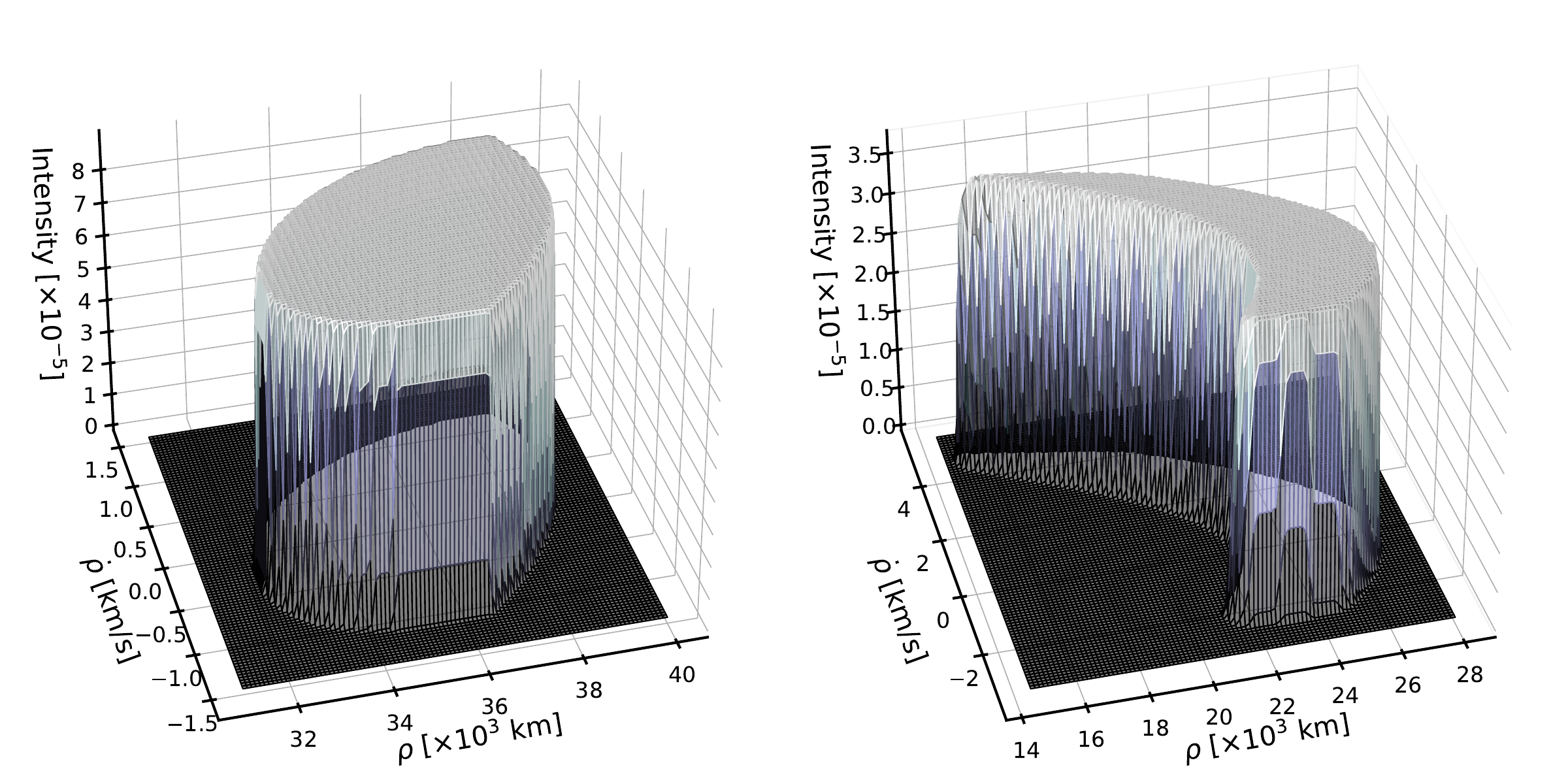}
\caption{\ac{GMM} approximation of the intersection of the constraint sets for the \ac{GEO} (left) and \ac{GTO} (right) test cases, described in the results section.}\label{fig:AR_GMM}
\end{figure}

In Fig. \ref{fig:AR_figure} we show the \ac{AR} constraints generated by the two numerical test cases used later in this paper -- the left being that generated for a \ac{GEO} target and the right for a \ac{GTO}. The overlap between the constraint boundaries constitutes the feasible range and range rate domain of a target. For the \ac{GEO} case, one can note the significant reduction in uncertainty produced by the eccentricity constraint. On the other hand, for the \ac{GTO} case, we do not impose such a tight constraint on eccentricity, so the feasible range/range rate space is much larger. In Fig. \ref{fig:AR_GMM} we show the \ac{GMM} approximation of the intersection of \ac{AR} constraints for the two cases. Using a sufficiently larger number of Gaussian components serves to well-approximate the uniform distribution between the \ac{AR} constraint boundaries.


\subsection{The \ac{CPHD} Filter}\label{sec:cphd_filter}
An \ac{RFS}-based description of the multi-target state provides a convenient formulation that allows us to consider the information contained in an empty measurement set, missed detections, false-alarms, and the presence of multiple unknown targets within a search domain. We denote the multi-target state and measurement spaces as $\mathcal{X} \subseteq \mathbb{R}^{n_x}$ and $\mathcal{Z} \subseteq \mathbb{R}^{n_z}$, respectively. The \ac{RFS} corresponding to the multi-target state and measurement at discrete time step $k$ are
\begin{equation}
    \boldsymbol{X}_k = \{\boldsymbol{x}_{k, 1}, \cdots, \boldsymbol{x}_{k, N_k} \} \in \mathcal{F}(\mathcal{X})\mbox{,}
\end{equation}
and,
\begin{equation}
    \boldsymbol{Z}_k = \{\boldsymbol{z}_{k, 1}, \cdots, \boldsymbol{z}_{k, M_k} \} \in \mathcal{F}(\mathcal{Z}){,}
\end{equation}
respectively. In the above, $N_k$ denotes the number of targets, and $M_k$ the number of measurements at the $k^{\text{th}}$ discrete step. $\mathcal{F}(\mathcal{X})$ and $\mathcal{F}(\mathcal{Z})$ are the collections of all finite subsets of $\mathcal{X}$ and $\mathcal{Z}$. The multi-target Bayes recursion is defined by the transition and update equations
\begin{equation}
    f_{k|k - 1}(\boldsymbol{X}_k|\boldsymbol{Z}_{1:k-1}) = \int \pi_{k|k-1}\left(\boldsymbol{X}_k|\boldsymbol{X}\right)f_{k - 1}(\boldsymbol{X}|\boldsymbol{Z}_{1:k-1}) \delta \boldsymbol{X}, \label{eqn:multi_target_transition} 
\end{equation}
and, 
\begin{equation}
    f_{k}(\boldsymbol{X}_k|\boldsymbol{Z}_{1:k}) = \frac{g_k(\boldsymbol{Z}_k|\boldsymbol{X}_k) f_{k|k - 1}(\boldsymbol{X}_k|\boldsymbol{Z}_{1:k -1})}{\int g_k( \boldsymbol{Z}_k|\boldsymbol{X}) f_{k|k-1}(\boldsymbol{X}|\boldsymbol{Z}_{1:k-1}) \delta \boldsymbol{X}} \mbox{,} \label{eqn:multi_target_update}
\end{equation}
where $\pi_{k|k - 1}(\boldsymbol{X}_k|\boldsymbol{X})$ and $g_k(\boldsymbol{Z}_k|\boldsymbol{X})$ are the multi-target transition kernel and likelihood function, respectively. In general, the recursion does not permit a tractable solution, so various approximations exist in the multi-target tracking literature; the simplest being the \ac{PHD} filter \cite{Mahler2003}. The \ac{PHD} filter assumes that the target number is Poisson distributed, and therefore tracks only the first-order moment of the multi-target state,
\begin{equation}
    D_{k}(\boldsymbol{x}| \boldsymbol{Z}_{1:k}) = \int  f_{k}\big(\boldsymbol{X}|\boldsymbol{Z}_{1:k} \big) \delta \boldsymbol{X}\label{eqn:intensity_function} \mbox{,}
\end{equation}
which is often referred to as the intensity function. The intensity function can be interpreted as a spatial density of targets over a search space. Because of its assumptions, the \ac{PHD} filter is highly sensitive to missed detections and clutter returns, as target cardinality information is inherently truncated. Furthermore, the \ac{PHD} filter assumptions impose that measurement clutter must be an independent and identically distributed (i.i.d.) Poisson process. This does not conform to scenarios in which the clutter intensity models spurious measurements arising from known background objects, such as ours.

To address the issues associated with the 
\ac{PHD} recursion, the \ac{CPHD} filter jointly tracks the intensity function, and a cardinality distribution describing the true target number \cite{Mahler2007}. In this case, the multi-target state is approximated by an i.i.d. cluster process \cite{Vo2007}. Additionally, the \ac{CPHD} assumes that each of the targets composing the multi-target state are mutually independent and independent with respect to measurement clutter, which is also an i.i.d. cluster process. The cardinality distribution is 
\begin{equation}
    \rho(n|\boldsymbol{Z}_{1:k}) = \frac{1}{n!} \int f_{k}\big(\{\boldsymbol{x}_1, \cdots, \boldsymbol{x}_n \}| \boldsymbol{Z}_{1:k}\big ) d\boldsymbol{x}_1 \cdots d\boldsymbol{x}_n\mbox{,}
\end{equation}
and together with the intensity function, satisfy the property
\begin{equation}
    \hat{N}_k = \sum_{n = 0}^\infty n\rho(n|\boldsymbol{Z}_{1:k}) = \int D_{k}(\boldsymbol{x}|\boldsymbol{Z}_{1:k}) d\boldsymbol{x}\mbox{.}
\end{equation}
In other words, the expected number of targets within the search space, $\hat{N}_k$, is the total integral of the intensity function. We use a \ac{GMM} parameterization of the intensity function for accurately representing the density attributed to the multi-target state,
\begin{equation}
    D_{k}(\boldsymbol{x}) = \sum_{i = 1}^{N_\text{mix}} w_i \mathcal{N}(\boldsymbol{x}; \boldsymbol{\mu}_i, P_i)\mbox{,}
\end{equation}
where $N_{\text{mix}}$ is the number of mixture components composing the \ac{GMM}, and $\mathcal{N}(\cdot; \boldsymbol{\mu}_i, P_i)$ is a Gaussian distribution parameterized by mean $\boldsymbol{\mu}_i$ and covariance $P_i$. This \ac{GMM} representation however is only one choice, and in fact, to provide a tractable information theoretic objective function for closed-loop sensor steering, we use an intermediate representation based on particles, which is further described in a later section of this work. In the following, we provide the general form of the \ac{CPHD} recursion, but refer the reader to \cite{Ristic2012} and \cite{Vo2007} as helpful references to understand the nuances of the \ac{GMM} and particle forms. 

%
%
%
The \ac{CPHD} recursion is composed of a prediction and update step. For our application, once the search set associated to a novel detection has been initialized, we impose that there are no additional target births and no target deaths in the prediction step of the recursion. The \textit{a priori} predicted intensity and cardinality functions are then
\begin{equation}
    D_{k|k - 1}(\boldsymbol{x}) = \int \pi_{k|k-1}(\boldsymbol{x}|\boldsymbol{\zeta})D_k(\boldsymbol{\zeta}) d\boldsymbol{\zeta}\label{eqn:intensity_transition_density}\mbox{,} 
\end{equation}
and, 
\begin{equation}
    \rho_{k|k-1}(n) = \rho_{k}(n)\mbox{,}
\end{equation}
respectively, where here $\pi_{k|k-1}(\boldsymbol{x}| \cdot)$ is the single-target transition kernel between discrete time steps $k - 1$ and $k$. The \ac{CPHD} update equations are significantly more complicated than those of the \ac{PHD} filter. For the intensity and cardinality functions, these are 
\begin{align}
    D_k(\boldsymbol{x}) &= \frac{\langle \Upsilon^1_k[D_{k|k-1}; \boldsymbol{Z}_k ], \rho_{k|k-1} \rangle}{\langle \Upsilon^0_k[D_{k|k-1}; \boldsymbol{Z}_k ], \rho_{k|k-1} \rangle} [1 - p_{D, k}(\boldsymbol{x})]D_{k|k - 1}(\boldsymbol{x}) \nonumber \\ 
    &+\sum_{\boldsymbol{z}\in \boldsymbol{Z}_k} \frac{\langle \Upsilon^1_k[D_{k|k-1}; \boldsymbol{Z}_k \backslash \{\boldsymbol{z}\} ], \rho_{k|k-1} \rangle}{\langle \Upsilon^0_k[D_{k|k-1}; \boldsymbol{Z}_k ], \rho_{k|k-1} \rangle} \psi_{k, \boldsymbol{z}} D_{k|k - 1}(\boldsymbol{x})\label{eqn:CPHD_intensity_update}
\end{align}
and,
\begin{equation}
      \rho_k(n) = \frac{\Upsilon^0_k[D_{k|k-1}; \boldsymbol{Z}_k ](n) \rho_{k|k-1}(n)}{\langle \Upsilon^0_k[D_{k|k-1}; \boldsymbol{Z}_k ], \rho_{k|k-1} \rangle }\mbox{,}\label{eqn:CPHD_cardinality_update}
\end{equation}
respectively. Here we use brackets, $\langle \cdot, \cdot \rangle$, to denote the inner products of two functions or vectors. In the above, $p_{D, k}(\boldsymbol{x})$ is the detection probability function, given as 
\begin{equation}
    p_{D, k}(\boldsymbol{x}) =
    \begin{cases}
        P_D, & \text{if}\ \boldsymbol{x} \in \mathcal{S}(\boldsymbol{u}_k; \boldsymbol{p})\mbox{,}\\
        0, & \text{otherwise}\mbox{,}\label{eqn:detection_probability_function}
    \end{cases}
\end{equation}
where $\mathcal{S}(\boldsymbol{u}_k; \boldsymbol{p})$ is the subset of the single target state space detectable by the sensor, conditioned on the $k^{\text{th}}$ action of the sensor, $\boldsymbol{u}_k$, and the sensor configuration parameters, $\boldsymbol{p}$. The term
\begin{equation}
    \psi_{k, \boldsymbol{z}}(\boldsymbol{x}) = \frac{\langle 1, \kappa_k \rangle}{\kappa_k(\boldsymbol{z})} g_k(\boldsymbol{z}| \boldsymbol{x}) p_{D, k}(\boldsymbol{x})\mbox{,}
\end{equation}
where $\kappa_k(\boldsymbol{x})$ is the spatial clutter intensity function that, for our application, models a known catalog of background objects. The function $g_k(\cdot|\boldsymbol{x})$ is the single target likelihood function. The constituent terms in Eq. \ref{eqn:CPHD_intensity_update} and Eq. \ref{eqn:CPHD_cardinality_update} are defined as
\begin{align}
    \Upsilon_k^u[D_{k|k-1}; \boldsymbol{Z}_k](n) &= \sum_{j = 0}^{\text{min}(|\boldsymbol{Z}_k|, n)}(|\boldsymbol{Z}_k| - j)!\rho_{K, k}(|\boldsymbol{Z}_k| - j)P_{j + u}^n\times \nonumber \\ 
    &\frac{\langle 1 - p_{D, k}, D_{k|k - 1} \rangle^{n - (j + u)}}{\langle 1, D_{k|k-1} \rangle^n} e_j \left(\boldsymbol{\Xi}(D_{k|k-1}, \boldsymbol{Z}_k)\right)\mbox{,}
\end{align}
where,
\begin{equation}
    \boldsymbol{\Xi}(D_{k|k-1}, \boldsymbol{Z}_k) = \{\langle D_{k|k-1}, \psi_{k, \boldsymbol{z}} \rangle : \boldsymbol{z} \in \boldsymbol{Z}_k \}\mbox{,}
\end{equation}
and $P_j^n$ is the permutation coefficient $\frac{n!}{(n - j)!}$. The elementary symmetric functions $e_j(\cdot)$ are defined as 
\begin{equation}
    e_j(\boldsymbol{W}) = \sum_{\boldsymbol{S}\subseteq \boldsymbol{W}, |\boldsymbol{S}| =j} \left( \prod_{\boldsymbol{\zeta} \in \boldsymbol{S}} \boldsymbol{\zeta} \right)\mbox{,}
\end{equation}
and can be computed recursively for each $j$ using a polynomial expansion with $e_0(\boldsymbol{W}) = 1$ by convention \cite{Assche1996}.

\subsection{Vetting Clutter and Target Measurements}\label{sec:clutter_model}

When a follow-up sensor scans the search set, it is possible that catalog space objects generate additional measurements. Typical strategies that take clutter as a spatially uniform Poisson process are effective for erroneous measurements arising from shot noise and other transient sensor responses. However, these strategies may produce target false-positives in the presence of persistent returns arising from catalog objects. Here, following that originally presented in \cite{Gehly2018} for tracking \ac{GEO} targets, we develop a clutter model based on an i.i.d cluster process that limits false-positives due to these objects. The spatial intensity function of clutter returns in the \ac{FOR} is represented by a \ac{GMM} so that 
\begin{equation}
    \kappa_k(\boldsymbol{z})  = \sum_{i = 1}^{N_{\kappa}} w_{\kappa, k}^i \mathcal{N}(\boldsymbol{z}; \boldsymbol{z}_{\kappa, k}^i, S_{\kappa, k}^i)\mbox{,}
\end{equation}
where, 
\begin{equation}
    \boldsymbol{z}_{\kappa, k}^i = \boldsymbol{h}(\boldsymbol{\mu}_{\kappa,k}^i; \boldsymbol{p})\mbox{,} \hspace{3ex} S_\kappa^i = H P_{\kappa, k}^i H^\top + R.\label{eqn:clutter_intensity_eqn}
\end{equation}
In the above, $w_{\kappa, k}^i$ is the weight attributed to the $i^{\text{th}}$ mixand composing the \ac{GMM} clutter intensity function. The variables $\boldsymbol{\mu}^i_{\kappa, k}$ and $P^i_{\kappa, k}$ are the predicted Cartesian state mean and covariance of the $i^{\text{th}}$ catalog mixand at time step $k$, respectively,  and $R$ is the measurement noise covariance matrix for the sensor. 

While not necessary, we assume that the uncertainty attributed to the clutter intensity is well constrained, so each catalog object is represented by a single Gaussian mixture component. Therefore, $w_{\kappa, k}^i = 1,\ \forall i = \{1, \cdots, n_{\kappa}\}$, where $N_{\kappa}$ is the number of catalog objects. Following this, we can represent the cardinality distribution of clutter measurements as
\begin{equation}
    \rho_{\kappa, k}(n) = \binom{N_{\kappa}}{n} P_D^n (1 - P_D)^{N_{\kappa} - n}\mbox{.}\label{eqn:clutter_cardinality_eqn}
\end{equation}
Here, $N_{\kappa}$ is the number of binomial trials. The success probability of each trial is taken to be the probability of detecting the catalog object, $P_D$. 


\subsection{\ac{GMM} Splitting for \ac{FOV} Partitioning}\label{sec:splitting}

In this section, we detail our \ac{GMM} splitting procedure that accounts for the discontinuity in detection probability between portions inside and outside the \ac{FOV} of the follow-up sensor. Our method shares commonalities with that in \cite{LeGrand2022}. There, the authors develop a method for a \ac{GMM} splitting when the measurement is a subset of the state space. Our method extends this to cases in which there exists a nonlinear map between the state and measurement spaces. Additionally, the authors in \cite{LeGrand2022} define a splitting criterion based on the statistical overlap of a mixand with the \ac{FOV} using a set of collocation points generated by a Gaussian distribution. We find it more convenient for our application to base the splitting criterion on a statistical distance of the mixand to the nearest edge of the \ac{FOV}.

As noted previously, we use a \ac{GMM} parameterization of the \ac{CPHD} intensity function to accurately represent the multi-target state in the filter recursion. The \ac{GMM} serves as the search set for follow-up sensor scans, however, when projected onto the \ac{FOR}, is typically much larger than the sensor's finite \ac{FOV}. Because Eq. \ref{eqn:CPHD_intensity_update} and Eq. \ref{eqn:CPHD_cardinality_update} are functionally dependent on the detection probability, to correctly update the \ac{CPHD}, one must partition the prior intensity into portions that are detectable and undetectable when conditioned on a sensor's pointing direction. 

Mixture components that overlap with the \ac{FOV} in a statistical distance-sense must be split to accurately generate this partition. Consider for now a single Gaussian mixand composing the prior intensity with weight $w$, mean $\boldsymbol{\mu}$, and covariance $P$. The mixand's projected mean and covariance onto the \ac{FOR} are
\begin{equation}
    \Tilde{\boldsymbol{\mu}} = \boldsymbol{h}(\boldsymbol{\mu}; \boldsymbol{p}, t)
\end{equation}
and,
\begin{equation}
    \Tilde{P} = H P H^\top\mbox{,}\label{eqn:projected_covariance}
\end{equation}
respectively. The function $\boldsymbol{h}(\cdot)$ is the map from the Cartesian state to an angular bearing measurement, and 
\begin{equation}
    H = \frac{\partial \boldsymbol{h}(\boldsymbol{x}; \boldsymbol{p}, t)}{\partial \boldsymbol{x}}\bigg|_{\boldsymbol{\mu}}\mbox{.}
\end{equation}
We use a criterion for splitting based on the Mahalanobis distance of the projected mixand to the nearest edge of a rectilinear \ac{FOV}, defined by a set of vertices, $\{\boldsymbol{a}_i\}_{i = 1}^4 $. Each vertex is transformed into a new coordinate system through a whitening transformation,
\begin{equation}
    \boldsymbol{b}_i = \Tilde{\Lambda}^{-1/2} \Tilde{V}^\top(\boldsymbol{a}_i - \Tilde{\boldsymbol{\mu}})\mbox{,}
\end{equation}
where the matrices $\Tilde{\Lambda}$ and $\Tilde{V}$ are determined via an eigendecomposition of the projected mixand, 
\begin{equation}
    \Tilde{V} \Tilde{\Lambda} \Tilde{V}^\top = \Tilde{P}\mbox{.} \label{eqn:eigen_decomp_projected_covariance}
\end{equation}
In the new coordinate system, $\Tilde{\boldsymbol{\mu}}$ is the origin, and the Euclidean distance is equivalent to the Mahalanobis distance in the original coordinates. Let $\boldsymbol{b}_{ij} = \boldsymbol{b}_j - \boldsymbol{b}_i$ define the vector connecting two adjacent vertices of the transformed \ac{FOV}, then the point on the line that minimizes the Mahalanobis distance of the projected mixand to said line is determined by 
\begin{equation}
    \boldsymbol{o}_{ij} =
    \begin{cases}
      \boldsymbol{b}_i - \frac{\langle \boldsymbol{b}_i, \boldsymbol{b}_{ij} \rangle}{\langle \boldsymbol{b}_{ij}, \boldsymbol{b}_{ij}\rangle} \boldsymbol{b}_{ij}, & \text{if}\ \langle \boldsymbol{b}_i - \boldsymbol{o}_{ij}, \boldsymbol{b}_j - \boldsymbol{o}_{ij}\rangle \leq 0, \\
      \boldsymbol{b}_i, & \text{if}\ ||\boldsymbol{b}_i|| < ||\boldsymbol{b}_j||, \\
      \boldsymbol{b}_j, & \text{otherwise}\mbox{.}
    \end{cases}
\end{equation}
A condition for splitting is triggered when 
\begin{equation}
   \min_{\boldsymbol{b}_i \rightarrow \boldsymbol{b}_j} ||\boldsymbol{o}_{ij}|| \leq d_{\text{M}}\mbox{,}\label{eqn:condition_for_split}
\end{equation}
where we use the notation $\boldsymbol{b}_i \rightarrow \boldsymbol{b}_j$ to denote the connection between any two adjacent vertices circumscribing the transformed \ac{FOV}. We define $d_{\text{M}}$ as a user-defined lower bound for an admissible Mahalanobis distance. Increasing the value of $d_{\text{M}}$ increases the boundary resolution of the partition, however, it also increases the resulting number of new Gaussian components. Therefore, a trade-off exists between accuracy and computational efficiency.

The next logical question to ask is how to perform the split. In this work, we split a \ac{GMM} component in the Cartesian state space along an optimal principal axis direction of $P$ \cite{DeMars2013b}. The resulting Gaussian mixture approximates the original mixand with a number of components that have a covariance smaller than the original. To determine the new components, we first form a univariate standard normal split library by minimizing 
\begin{equation}
    J = L_2(q||\hat{q}) + \lambda \hat{\sigma} \hspace{3ex} \text{s.t.} \hspace{0.5ex} \sum_{i = 1}^{N_\text{split}} \hat{\alpha}_i = 1\mbox{,}
\end{equation}
where $L_2(\cdot)$ is the $\beta$-divergence between two distributions with the parameter $\beta = 2$ and $\lambda$ is a regularization term that controls the size of the resulting split mixture component's covariance. The probability distribution $q = \mathcal{N}(x; 0, 1)$, and 
\begin{equation}
    \hat{q} = \sum_{i = 1}^{N_{\text{split}}} \hat{\alpha}_i \mathcal{N}(x; \hat{m}_i, \hat{\sigma}_i)\mbox{,}
\end{equation}
where $N_{\text{split}}$ is the resulting component number. In practice, the univariate split library is computed off-line, and we list the resulting univariate mixture weights, $\hat{\alpha}_i$, means, $\hat{m}_i$, and standard deviations, $\hat{\sigma}_i$, for the three-component split library used in this work in Table \ref{tab:split_library}. 
\begin{table}[h]
\caption{Three-component univariate split library with $\beta = 2$ and $\lambda = 0.001$ provided by \cite{DeMars2013b}.}\label{tab:split_library}%
\begin{tabular}{ccccccc}
\toprule
$i$ & $\hat{\alpha}$  && $\hat{m}_i$ && $\hat{\sigma}_i$ \\
\midrule
1    & 0.2252246249   && -1.0575154615  && 0.6715662887  \\
2    & 0.5495507502   && 0  && 0.6715662887  \\
3    & 0.2252246249   && 1.0575154615  && 0.6715662887  \\
\botrule
\end{tabular}
\end{table}
The univariate split is applied to an optimal principal axis, indexed by $\ell^*$, of the original mixand so that new mixture components are parameterized by the following:
\begin{equation}
    \hat{w}_i = \hat{\alpha}_i w, \hspace{3ex} \hat{\boldsymbol{\mu}}_i = \boldsymbol{\mu} + \sqrt{\lambda_\ell} \hat{m}_i \boldsymbol{v}_k, \hspace{3ex} \hat{P}_i = V \Lambda_i V^\top\mbox{,}
\end{equation}
where $\boldsymbol{v}_{\ell^*}$ is the $\ell^{* \text{th}}$ eigenvector of an eigendecomposition of $P$, and 
\begin{equation}
    \Lambda_i = \text{diag}\left( [\lambda_1, \cdots, \hat{\sigma}_i^2 \lambda_{\ell^*}, \cdots, \lambda_n] \right)\mbox{.}
\end{equation}

\ac{GMM} splitting is performed recursively so that each original component is tested using Eq. \ref{eqn:condition_for_split}, and the resulting split mixtures are re-tested and split until the accepted Mahalanobis distance criterion is met. It is desirable to minimize the final number of mixture components to aid in the computational tractability of the \ac{CPHD} recursion. Critical to this is choosing the split direction, and we do this by choosing the direction that exhibits the greatest sensitivity to the largest principal direction of its projection onto the \ac{FOR}. To determine this, using Eq. \ref{eqn:projected_covariance} and Eq. \ref{eqn:eigen_decomp_projected_covariance}, we can first create the following equivalence relation:
\begin{equation}
    \Tilde{V} \Tilde{\Lambda} \Tilde{V}^\top =  H V \Lambda V^\top H^\top \mbox{,}
\end{equation}
and rearranging,
\begin{equation}
    \Tilde{\Lambda} = \Tilde{V}^\top H V \Lambda^{1/2} \Lambda^{1/2} V^\top H^\top \Tilde{V}\mbox{.}\label{eqn:rearranged_equivalence_relation}
\end{equation}
The matrix $\Tilde{\Lambda}$ is a diagonal matrix containing eigenvalues of the projected mixand. Using Eq. \ref{eqn:rearranged_equivalence_relation}, we can see the eigenpair of $P$ that has the largest sensitivity to $\Tilde{\lambda}_{\text{max}}$ is
\begin{equation}
    \boldsymbol{v}^*, \lambda^* = \argmax_{\boldsymbol{v}_i, \lambda_i} | \sqrt{\lambda_i} \langle \Tilde{\boldsymbol{v}}_{\text{max}},  H \boldsymbol{v}_i \rangle |\mbox{,}
\end{equation}
where $\Tilde{\boldsymbol{v}}_{\text{max}}$ is the eigenvector corresponding to $\Tilde{\lambda}_{\text{max}}$.

\begin{figure}[hbt!]
\centering
\includegraphics[width=1.0\textwidth]{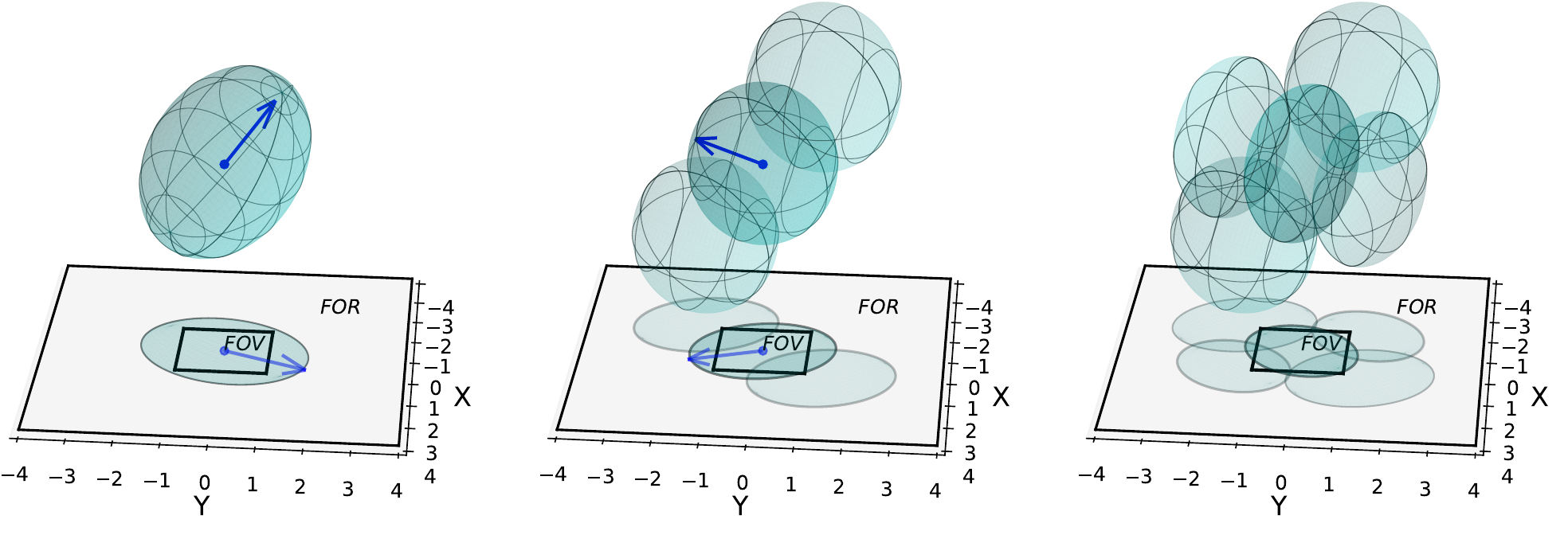}
\caption{Illustration for the recursive splitting operation used in this work. Here we show the original mixand and its resulting mixture after one and two splits.}\label{fig:splitting_figure}
\end{figure}

Splitting is performed in the 6D Cartesian state space of the space object, but to illustrate this operation here we quickly summarize a 3D linear example. In this test case, a \ac{GMM} is composed of a single mixand with 
\begin{equation}
    w = 1.0,\ \hspace{1ex} \boldsymbol{\mu} = [0, 0, 3]^\top,\ \hspace{1ex}  P = \begin{bmatrix}
        2.5,& 0.5& 1.5\\
        0.5,& 3.5,& 1\\
        1.5,& 1,& 3
    \end{bmatrix}\mbox{,}
\end{equation}
and, 
\begin{equation}
    \boldsymbol{h}(\boldsymbol{x}) = [x_1, x_2]^\top,\ \hspace{1ex} H = \begin{bmatrix}
        1,& 0,& 0\\
        0,& 1,& 0
    \end{bmatrix}\mbox{.}
\end{equation}
In Fig. \ref{fig:splitting_figure}, from left to right, we show the $1\sigma$ covariance ellipsoid of the original mixand and its projection into the \ac{FOR}, and those after one and two recursive splits, respectively. The blue arrows in the left two figures indicate the split direction in the 3D space and the corresponding maximum eigen-direction in the \ac{FOR}. The \ac{FOV} for this simple case is a 2D rectilinear box centered at the origin with a side length of two.


\subsection{Information-Based Sensor Control}\label{sec:sensor_control}

This section describes an approach for directing follow-up sensor scans and extends our work presented in \cite{Wolf2023}. This problem can be posed as a \ac{POMDP} in which a tasking agent's actions encode the angular pointing direction of the optical telescope and the information state is the \ac{CPHD} associated to the untracked targets. The agent seeks to maximize its reward that we take to be a function that quantifies the expected information gain generated through a specified action. Information theoretic sensor tasking approaches have found broad applicability in the \ac{SDA} literature (e.g., \cite{Sunberg2016, Gehly2018, Adurthi2020}), however, almost exclusively rely on the \ac{PIMS} approximation \cite{Mahler2003}. This, however, does not account for expected missed detections and false-positive detections in the information gain functional. Most importantly to our application, the \ac{PIMS} fails to consider the expected information gain contained in an empty measurement set, colloquially termed \textit{negative information}. 

The authors in \cite{LeGrand2021} address these known issues by optimizing sensor actions with respect to a form of the \ac{KLD} based on a \ac{cell-MB} approximation. They apply their approach to optimally directing a sensor to seek and track terrestrial targets within a 2D scene using a multi-Bernoulli filter. In this work, we take an alternative approach compatible with our \ac{CPHD} description of the multi-target state. We derive a form of the expected R\'enyi-divergence between the prior and posterior \ac{CPHD} based on weighted particle nearest-neighbors distances. This expected divergence approximation provides a computationally tractable reward used in closed-loop sensor control. 

For two arbitrary \ac{PDF}s, $p(\boldsymbol{x})$ and $q(\boldsymbol{x})$, the R\'enyi-divergence, also referred to as the $\alpha$-divergence, is 
\begin{equation}
    \mathcal{D}_{\alpha}(p(\boldsymbol{x})||q(\boldsymbol{x})) = \frac{1}{\alpha - 1} \log \int p^\alpha(\boldsymbol{x}) q^{1 - \alpha}(\boldsymbol{x}) d\boldsymbol{x} \mbox{,}\label{eqn:renyi_divergence}
\end{equation}
where $\alpha \in (0, 1) \cup (1, \infty)$ and the distribution $q(\boldsymbol{x})$ has full support. The R\'enyi-divergence can loosely be thought as a measure of dissimilarity between probability densities, in which the parameter $\alpha$ controls the discriminating power to the tails of the distributions. It shares many of the same properties as the \ac{KLD}, and in fact, in the limit as $\alpha \rightarrow 1$, the R\'enyi-divergence is equivalent to the \ac{KLD} \cite{vanErven2014}.  Following that presented in \cite{Ristic2011}, the definition in Eq. \ref{eqn:renyi_divergence} can be extended to a multi-target prior and posterior so that
\begin{equation}
    \mathcal{D}_{\alpha}(f_{k|k}||f_{k|k-1}) = \frac{1}{\alpha - 1} \log \int \left[f_{k|k}(\boldsymbol{X}_k|\boldsymbol{Z}_{1:k})\right]^\alpha \left[ f_{k|k-1}(\boldsymbol{X}_k|\boldsymbol{Z}_{1:k-1}\right]^{1 - \alpha} \delta \boldsymbol{X}\mbox{.}\label{eqn:multi_target_renyi}
\end{equation}
The intuition for its use in this context lies in that a posterior density will always contain the same or greater information content than the prior. Therefore, it is most advantageous to maximize the ``dissimilarity" between the two if we desire to maximize the information gain. As noted previously, the \ac{CPHD} is described by an i.i.d cluster process, where the multi-object state can be written as  
\begin{equation}
    f(\boldsymbol{X}) = n!\rho(n) \prod_{\boldsymbol{x}\in\boldsymbol{X}} s(\boldsymbol{x}).\label{eqn:cluster_RFS}
\end{equation}
In the above, $s(\boldsymbol{x})$ is the unit-normalized \ac{CPHD} intensity, given as 
\begin{equation}
    s(\boldsymbol{x}) = \frac{1}{\hat{N}} D(\boldsymbol{x})\mbox{.}
\end{equation}
For a steerable sensor, the $k^{\text{th}}$ measurement realization, $\boldsymbol{Z}_k$, is dependent on the action $\boldsymbol{u}_k$. Therefore, substituting Eq. \ref{eqn:cluster_RFS} into the multi-target R\'enyi-divergence definition generates our reward function, 
\begin{align}
    \mathcal{R}\left(\boldsymbol{u}_k, f_{k|k-1}(\boldsymbol{X}_k|\boldsymbol{Z}_{1:k-1}), Z_k(\boldsymbol{u_k})\right) &= \frac{1}{\alpha - 1} \log \sum_{n \geq 0} \rho_{k}(n; \boldsymbol{u}_k)^\alpha \rho_{k|k - 1}(n)^{1 - \alpha} \nonumber \\
    & \cdot \left [ \int s_{k}(\boldsymbol{x}; \boldsymbol{u}_k)^\alpha s_{k|k-1}(\boldsymbol{x})^{1-\alpha} d \boldsymbol{x} \right] \label{eqn:alpha_div_CPHD}\mbox{,}
\end{align}
which, from herein, we will refer to as $\mathcal{R}(\boldsymbol{u}_k)$. As noted in \cite{Hero2008}, the choice of $\alpha$ impacts the performance of a sensor management scheme. In cases where two distributions are quite similar, the authors suggest that $\alpha = 0.5$, equivalent to the Hellinger affinity, provides the maximum differentiation, and we adopt this recommendation in this work. While the reward function is dependent on the $k^\text{th}$ sensor action, the outcome of the action is non-deterministic. Deciding an optimal action is then determined by 
\begin{equation}
    \boldsymbol{u}_k^* = \argmax_{\boldsymbol{u}_k \in \mathcal{U}_k} \mathbb{E}\left[ \mathcal{R}(\boldsymbol{u}_k)\right]\mbox{,}\label{eqn:reward_function}
\end{equation}
where $\mathbb{E}[\cdot]$ denotes the statistical expectation taken over the measurement set, $\boldsymbol{Z}_k(\boldsymbol{u}_k)$, and $\mathcal{U}_k$ is the set of admissible sensor actions at time step $k$. 

The main challenge that we address is computing the integral term contained in Eq. \ref{eqn:alpha_div_CPHD}. One strategy employed in \cite{Gehly2018} is to evaluate the integral by directly replacing $s_{k|k-1}(\boldsymbol{x})$ and $s_{k}(\boldsymbol{x})$ with the unit-normalized prior and posterior \ac{GMM} representing the \ac{CPHD} intensity. This, however, does not admit an analytical solution, and while approximations to such integrals exist (e.g., \cite{Hershey2007, Huber2008}), they are generally computationally expensive for large mixtures. A \ac{GMM}-based approach is even more problematic for our problem because in deciding the optimal action in Eq. \ref{eqn:reward_function}, one must evaluate the expected impact of each potential action, which in itself, requires the \ac{GMM} splitting operation introduced above -- a computationally demanding task when repeated over all admissible actions in the set $\mathcal{U}_k$. 

To address these limitations, we use an intermediate representation of the \ac{CPHD} based on a particle distribution. First, let us define a parameterization so that
\begin{equation}
    s_{k|k-1}(\boldsymbol{x}) = \sum_{i = 1}^{N_{\text{samp}}} w_{k|k-1}^i\delta(\boldsymbol{x} - \boldsymbol{x}_{k|k-1}^i)\mbox{,}\label{eqn:particle_prior}
\end{equation}
and, 
\begin{equation}
    s_{k}(\boldsymbol{x}) = \sum_{i = 1}^{N_\text{samp}} w_{k}^i \delta(\boldsymbol{x} - \boldsymbol{x}_{k|k-1}^i)\mbox{,}\label{eqn:particle_posterior}
\end{equation}
where $N_{\text{samp}}$ is the number of importance samples, $w_{k|k-1}^i = 1/N_{\text{samp}}\ \forall i\in\{1, \cdots, N_{\text{samp}}\}$, and the set $\{w_k^i\}_{i = 1}^{N_\text{samp}}$ are the posterior weights produced by the particle \ac{CPHD} update (see \cite{Ristic2012} for implementation details). The function $\delta(\cdot)$ is the Dirac delta function. The particle states, $\{\boldsymbol{x}_{k|k - 1}^i\}_{i = 1}^{N_{\text{samp}}}$, are distributed according to the predicted \ac{GMM} intensity function, $D_{k|k-1}(\boldsymbol{x})$. This intermediate representation solves both of our challenges: 1) it produces a tractable closed-form expression for the integral term in Eq. \ref{eqn:alpha_div_CPHD}, and 2) this representation does not require mixture splitting as particles cannot span across the \ac{FOV} boundary. However, computing the integral term as written presumes $s_{k|k - 1}(\boldsymbol{x})$ and $s_k(\boldsymbol{x})$ to be discrete densities. It should be clear that this would be entirely inaccurate, as it neglects the measure's dependence on the spatial spread of particles. The authors in \cite{Skoglar2009} noted this and derive a form of the differential entropy using the prior and posterior densities in a particle filter. Their approximation, however, still relies on the ideal measurements.

\begin{figure}[hbt!]
\centering
\includegraphics[width=0.9\textwidth]{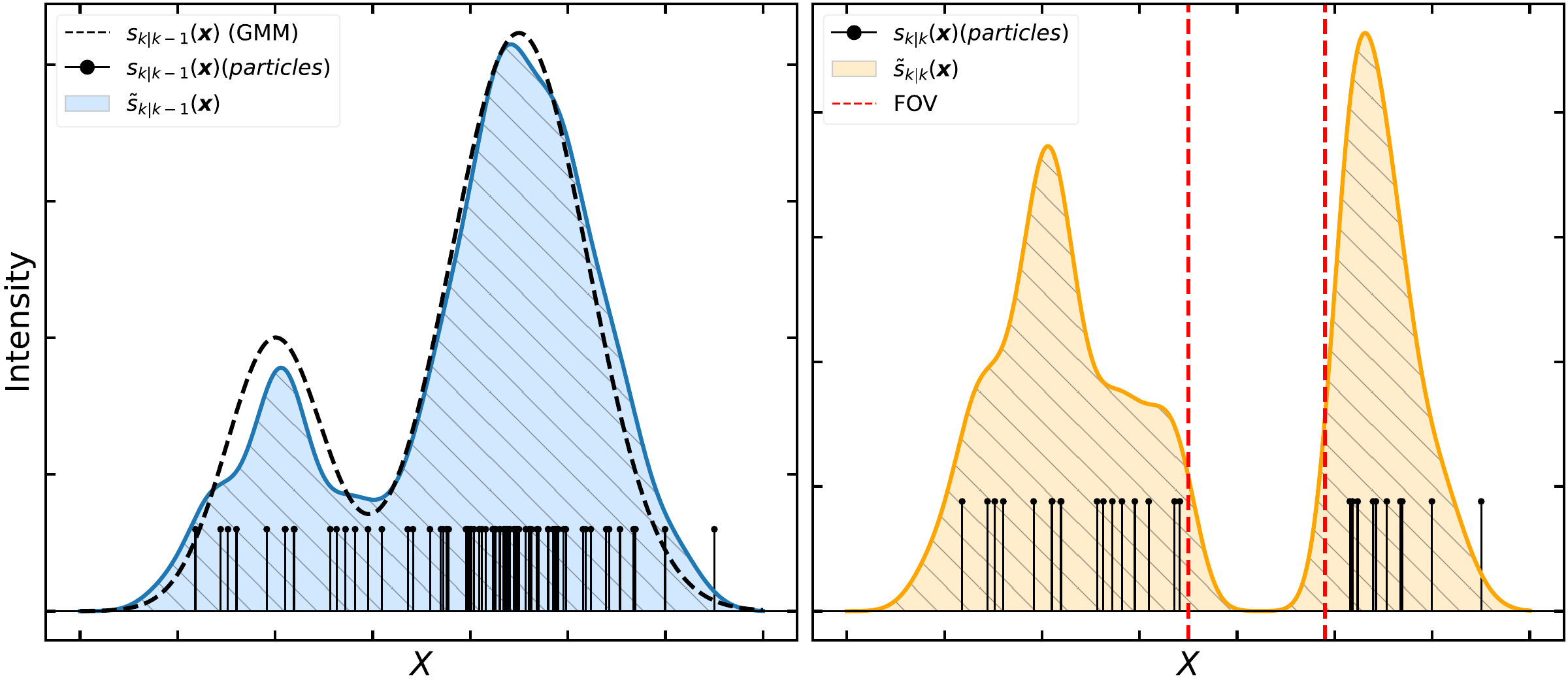}
\caption{Here, we illustrate the plug-in kernel intensity estimate in a 1D example for the prior (left) and posterior, given an empty measurement (right). These are used to generate the R\'enyi-divergence estimate in Eq. \ref{eqn:alpha_div_CPHD_particle_approx}.} \label{fig:kernel_figure}
\end{figure}

As suggested by \cite{Leoneko2008} in application to entropy estimates, and later generalized for weighted particles \cite{Jiri2011}, we replace the Dirac delta function in Eq.~\ref{eqn:particle_prior} and Eq.~\ref{eqn:particle_posterior} with a plug-in kernel estimate so that 
\begin{equation}
    \Tilde{s}_{k|k-1}(\boldsymbol{x}) = \sum_{i = 1}^{N_{\text{samp}}} \Tilde{w}_{k|k - 1}^{i}\frac{\Gamma(n_x/2 + 1)}{\varrho_i^{n_x} \pi^{n_x/2}} \delta_B(\boldsymbol{x} - \boldsymbol{x}^i)\mbox{,}\label{eqn:particle_prior_new}
\end{equation}
and,
\begin{equation}
    \Tilde{s}_k(\boldsymbol{x}) = \sum_{i = 1}^{N_{\text{samp}}} \Tilde{w}_k^i\frac{\Gamma(n_x/2 + 1)}{\varrho^{n_x}_i \pi^{n_x/2}}\delta_B(\boldsymbol{x} - \boldsymbol{x}^i)\mbox{.}\label{eqn:particle_posterior_new}
\end{equation}
In the above, $\Gamma(\cdot)$ is the gamma function and $\varrho_i$ is the distance of the particle $\boldsymbol{x}^i$ to its $\ell^{\text{th}}$ nearest-neighbor, not including itself. The new prior weights are defined as $\Tilde{w}_{k|k-1}^i = \ell/N_{\text{samp}}\ \forall i \in \{1, \cdots, N_{\text{samp}}\}$, and 
\begin{equation}
    \delta_B(\boldsymbol{x} - \boldsymbol{x}^i) = \begin{cases}
        1,& \text{if}\ ||\boldsymbol{x} - \boldsymbol{x}^i|| < \varrho_i\\
        0,& \text{otherwise}\mbox{.}
    \end{cases}
\end{equation}
The new posterior weights are defined as 
\begin{equation}
    \Tilde{w}_k^i = \sum_{j\in \mathcal{L}_k^i} w_k^j\mbox{,}
\end{equation}
where $\mathcal{L}_k^i$ is the set of $\ell - 1$ indices associated to the particles closest to $\boldsymbol{x}_i$, including the particle itself. In Fig.~\ref{fig:kernel_figure}, we provide a simple 1D illustration of this approximation. In the left panel, the prior samples, $s_{k|k-1}(\boldsymbol{x})$, are drawn from the predicted intensity, and are used to form the kernel density estimate, $\Tilde{s}_{k|k-1}(\boldsymbol{x})$. The right side illustrates the posterior particle weights and kernel density following a negative measurement scan, with the \ac{FOV} indicated by the vertical dashed red lines. By plugging in Eq. \ref{eqn:particle_prior_new} and Eq. \ref{eqn:particle_posterior_new}, we find that Eq. \ref{eqn:alpha_div_CPHD} reduces to
\begin{align}
    \mathcal{R}(\boldsymbol{u}_k) &= \frac{1}{\alpha - 1} \log \sum_{n \geq 0} \rho_{k}(n; \boldsymbol{u}_k)^\alpha \rho_{k|k - 1}(n; \boldsymbol{u}_k)^{1 - \alpha} \nonumber \\
    & \cdot \left [ \sum_{i = 1}^{N_{\text{samp}}} \left (\frac{\sum_{j \in \mathcal{L}_k^i} w
    _k^j}{\ell/N_{\text{samp}}}\right)^\alpha  
    \right] \label{eqn:alpha_div_CPHD_particle_approx}\mbox{.}
\end{align}
The nearest-neighbor index used for generating the sets $\{\mathcal{L}_k^i\}_{i = 1}^{N_{\text{samp}}}$ need only be computed once per time set $k$. The number of nearest neighbors has an effect on the accuracy of the R\'enyi-divergence estimate and should depend on the state dimension, $n_x$. Through empirical trials, we find that $\ell = 10$ works well for the 6D space object Cartesian state-space. We emphasize again that this intermediate particle representation is solely used to generate the divergence approximation, and the particles and nearest-neighbor index are discarded after each iteration, rather than being propagated to the next time step and used in the \ac{CPHD} measurement update step.

The next challenge that we address is computing the expectation in Eq. \ref{eqn:reward_function}. 
While the \ac{PIMS} approximation computes this expectation by evaluating the effect of a singular ideal measurement set, $\bar{\boldsymbol{Z}}_k(\boldsymbol{u}_k)$, this is insufficient for our case, as it may not be well known whether the control input will result in a true-positive detection from the untracked targets. Instead, for each potential control input, we use a Monte-Carlo integration over its corresponding measurement set distribution. A set realization is composed of clutter arising from background objects and positive measurements from the targets we are attempting to acquire, so that at each time
\begin{equation}
    \boldsymbol{Z}_k(\boldsymbol{u}_k) = \boldsymbol{Z}_{\kappa, k}(\boldsymbol{u}_k) \cup \boldsymbol{Z}_{\text{targ}, k}(\boldsymbol{u}_k)\mbox{.} 
\end{equation}
To generate a sample realization, we first draw a measurement number from the respective source. For those arising from clutter, 
\begin{equation}
    |\boldsymbol{Z}_{\kappa, k}(\boldsymbol{u}_k)| \sim \binom{N_{\kappa_\text{FOV}, k}}{n} P_D^n (1 - P_D)^{N_{\kappa_\text{FOV}, k} - n}\mbox{.}
\end{equation}
Here, $|\cdot|$ returns the cardinality of a set and $N_{\kappa_{\text{FOV}}, k}$ is the number of catalog objects contained inside the \ac{FOV} at time step $k$. The number of positive returns from the targets are distributed according to a multi-Bernoulli distribution \cite{Mahler2014}. This is generated via the intermediate particle intensity function introduced above and can be written as
\begin{equation}
    |\boldsymbol{Z}_{\text{targ}, k}(\boldsymbol{u}_k)| \sim \left(\prod_{j = 1}^{J_{\text{FOV}}}(1 - q_{k|k-1}^j)\right)\cdot e_n\left(\left\{\frac{q_{k|k-1}^1}{1 - q_{k|k-1}^1}, \cdots, \frac{q_{k|k-1}^{J_{\text{FOV}}}}{1 - q_{k|k-1}^{J^\text{FOV}}} \right\}\right)\mbox{.}
\end{equation}
Here, $q_{k|k-1}^j = P_D (\hat{N}_{k|k-1}\cdot  w_{\text{FOV},k|k-1}^j)$, where $w_{\text{FOV}, k|k-1}^j$ is the unit-normalized weight of the $j^{\text{th}}$ particle contained interior to the \ac{FOV}. The functions $e_n(\cdot)$ once again denote the roots of the elementary symmetric polynomials of order $n$. Following this, a number of measurements equal to the cardinality sample are drawn from their respective intensity function to produce the sample measurement set realization. This is repeated for $N_{\text{trials}}$ to produce the collection $\{\boldsymbol{Z}^i_k(\boldsymbol{u}_k)\}_{i = 1}^{N_{\text{trials}}}$. The response to each measurement set is evaluated using Eq. \ref{eqn:alpha_div_CPHD_particle_approx} and the average response provides our approximation of Eq. \ref{eqn:reward_function}. This procedure is repeated for all $\boldsymbol{u}_k \in \mathcal{U}_k$ and the action corresponding to the maximum reward is executed.

%
\section{Algorithm Summary}\label{sec:closing_the_loop}

\begin{figure}[hbt!]
\centering
\includegraphics[width=1.0\textwidth]{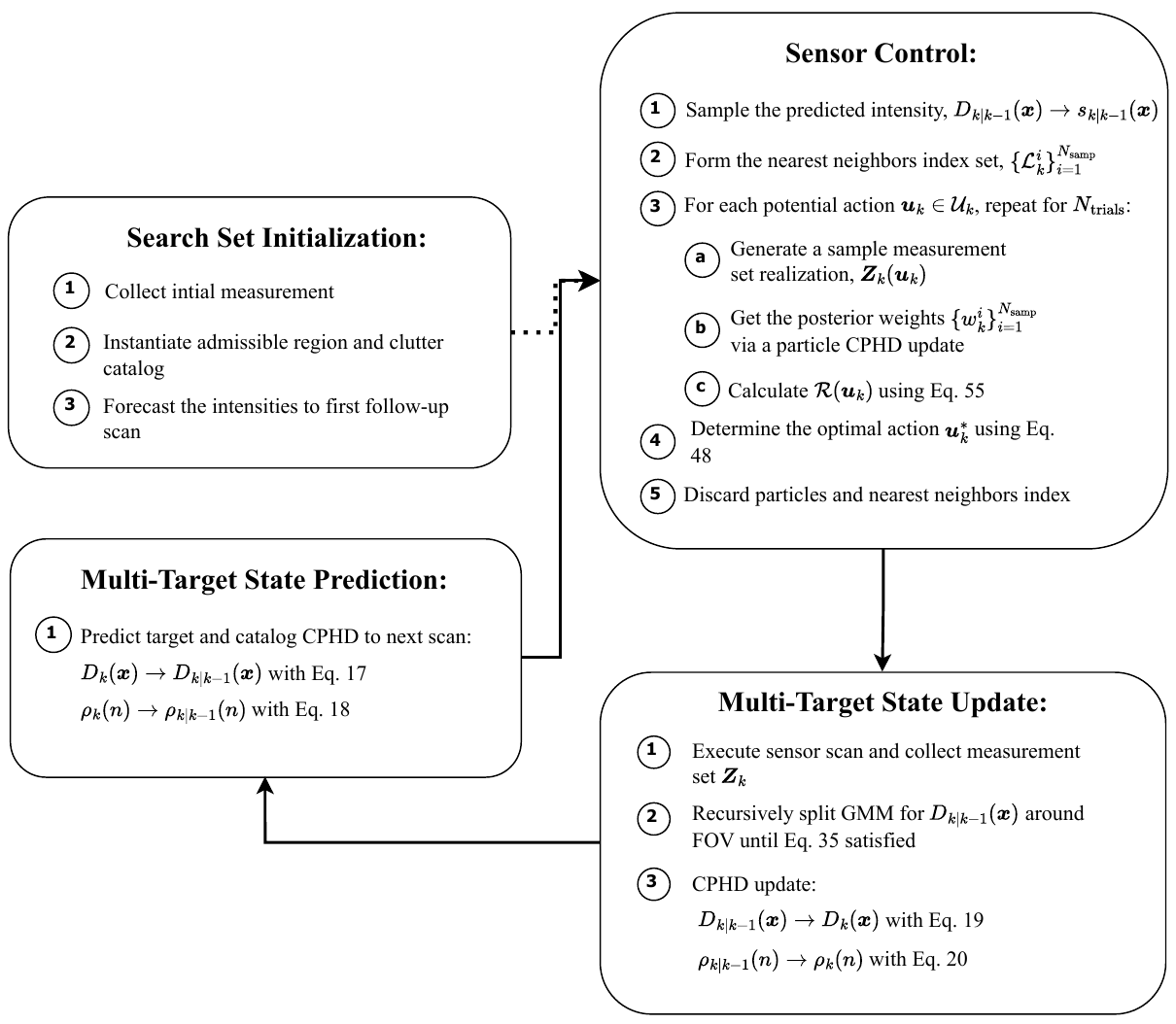}
\caption{Joint sensor control and multi-target filtering flowchart for seeking and tracking newly detected space objects.}\label{fig:flow_chart}
\end{figure}

Here, we briefly describe the synthesis of the tools previously introduced in generating a closed-loop sensor control and target tracking strategy for seeking and constraining novel space objects. Fig. \ref{fig:flow_chart} provides a flowchart for these operations. Given a novel detection, a search set is instantiated through the corresponding optical measurement and \ac{AR} constraints. The set is propagated to the starting epoch of the follow-up search, at which point, an optimal action, determined by our particle R\'enyi-divergence approximation, is selected. Upon this, the corresponding sensor scan is executed, and we apply a \ac{CPHD} update to the search set based on the collected measurement. The posterior search set, along with accompanying clutter catalog, is forecast to the next measurement epoch. These steps are recursively executed.

%
\section{Numerical Case Studies}\label{sec:results}

\subsection{Test Case Descriptions}

We present two numerical case studies demonstrating our methods for follow-up tasking over a search set generated via the \ac{AR} method. The first test case considers the acquisition of targets contained in a search set generated with a measurement of an object in a \ac{GTO} and the second for one in \ac{GEO}. To provide geographic realism for our examples, we use the catalog of primary \ac{SDA} sensor site locations contained in \cite{Choi2017}. In the first and second test cases, initial measurements are generated by an optical telescope located in Socorro, New Mexico and Diego Garcia, respectively. The follow-up search is conducted with optical telescopes located on the island of Maui and Ascension Island for those respective cases. The follow-up sensor's \ac{FOV} and measurement cadence is selected to be consistent with a fast focal ratio optical telescope capable a quickly scanning a search domain (e.g., the \ac{SST}\cite{darpa_sst}). However, these hardware specifications are not necessarily reflective of the sensing configurations that may be contained at the geographic locations used in this study. The $P_D = 0.75$, which is conservative, but serves to stress-test our methods, and in fact, may be characteristic of a scenario in which an actor is deliberately controlling the reflective surfaces of a space object for concealment \cite{Harrison2021}. Table \ref{tab:network_parameters} summarizes the sensor configuration parameters used for the initial detection and follow-up tasking for each of these cases.
\textbf{\begin{table}[h]
\caption{Sensor configuration parameters for each test case.}\label{tab:network_parameters}%
\begin{tabular}{cccccccccccc}
\toprule
Parameter && Case \#1 && Case \#2 \\
\midrule
Initial Sensor Mode && Optical  && Optical  \\
Follow-Up Sensor Mode && Optical  && Optical  \\
Initial Date-Time && 2024-04-09T05:00:00Z  && 2024-04-09T20:00:00Z  \\
$\Delta t$ Cut-Out (hours)   && 2.0  && 5.0  \\
Initial Sensor Lat./Long. (deg.)    && (34.0584, -106.8914)  && (-7.3195, 72.4229)  \\
Follow-Up Sensor Lat./Long. (deg.) && (20.7, -156.3)  && (-7.90663, -14.40258)  \\
Follow-Up Sensor \ac{FOV} (deg.) && 6.0  && 6.0  \\
Measurement Noise RMS (arcsec) && 3.0  && 3.0  \\
$P_D$ && 0.75 && 0.75 \\
Number Follow-Up Scans && 30  && 80  \\
$\Delta t$ Scan (sec.) && 15.0  && 15.0  \\
\botrule
\end{tabular}
\end{table}}

To generate the primary measurement, osculating Keplerian elements corresponding to each case are propagated forward until a feasible observation epoch is reached. These initial conditions are contained in Table \ref{tab:test_case_COEs}. We impose constraints on the viewing geometry such that a line-of-sight exists between the sensor and the target, the target is illuminated, and the sensor site is in shadow. Upon initial detection, a search set is generated using the \ac{AR} method as described previously. The semimajor axis and eccentricity bounds for the \ac{AR} constraints are contained in Table \ref{tab:AR_test_case_parameters}. A follow-up observation window for the secondary sensor is determined in a similar fashion.

\begin{table}[h]
\caption{Osculating Keplerian elements for each test case at the initial date-time.}\label{tab:test_case_COEs}%
\begin{tabular}{cccccccccccccc}
\toprule
Test Case && $a$ (km) && $e$ && $i$ (deg) && $\Omega$ (deg) &&  $\omega$ (deg) && $\nu$ (deg)\\ 
\midrule
Case \#1  && 25447.5 && 0.66 && 1.0 && 0.001 && 0.001 && 240.0 \\
Case \#2  && 42259.0 && 0.001 && 5.0 && 0.001 && 0.001 && 135.0 \\
\botrule
\end{tabular}
\end{table}
\begin{table}[h]
\caption{Parameters for initializing the search set in each test case.}\label{tab:AR_test_case_parameters}%
\begin{tabular}{cccccccccccc}
\toprule
Parameter &&&& Case \#1 &&&& Case \#2 \\
\midrule
Orbit Type &&&& GTO &&&& GEO\\
Propagator &&&& Vinti  &&&& Vinti  \\
Eccentricity Bounds   &&&& (0.0, 0.7)  &&&& (0.0, 0.35)  \\
Semimajor Axis Bounds ($\times 10^3$ km) &&&& (20.0, 42.0)  &&&& (10.0, 45.0)  \\
Target Number &&&& 2  &&&& 10  \\
Clutter Object Number &&&& 10  &&&& 15  \\
\botrule
\end{tabular}
\end{table}

\begin{figure}[hbt!]
\centering
\includegraphics[width=1.0\textwidth]{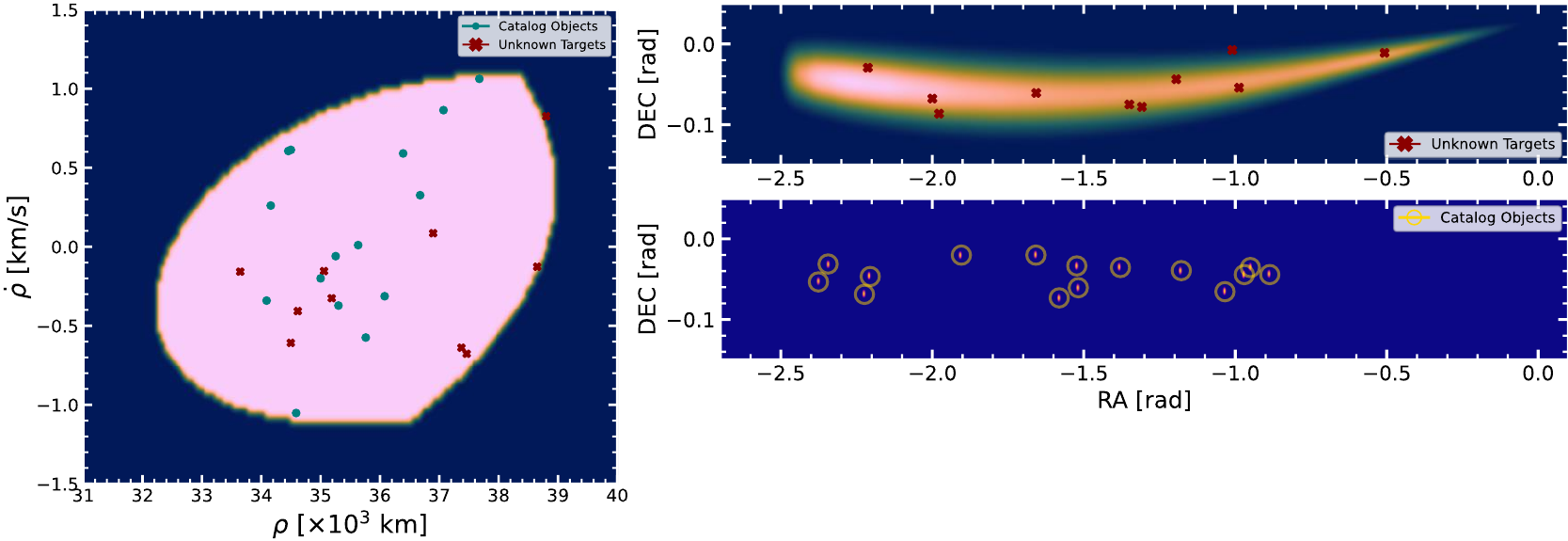}
\caption{On the left, we show the initial \ac{AR}, and sampled untracked target and clutter catalog objects for one Monte-Carlo realization, corresponding to test case \#2. The right forecasts the scenario to the initial follow-up measurement epoch, and shows the projected search set (top) and catalog object intensity (bottom).}
\label{fig:intial_targets_and_clutter}
\end{figure}

A set of target states are sampled from the \ac{GMM} representing the search set for a truth-model run. To mitigate any confusion, we note here that these target states do not necessarily correspond to the elements contained in Table \ref{tab:test_case_COEs}, but rather constitute a possible multi-target realization that we can use to evaluate our method. Similarly, we sample the \ac{AR} to produce dynamically consistent known objects from the same search domain. These are dubbed ``clutter catalog objects'' since the proposed \ac{CPHD} framework models them as clutter in the update.  This ensures that the catalog object states properly overlap so we can evaluate the clutter object vetting method presented in Section \ref{sec:clutter_model}. For each test case, we perform 100 Monte-Carlo trials, repeating this truth-model initialization process. In Fig. \ref{fig:intial_targets_and_clutter} we show the sampled targets and clutter catalog for one Monte-Carlo realization, corresponding to the case \#2. A Vinti analytic propagator, which includes Earth's $J_2$ and $J_3$ oblateness terms, generates the target and clutter states at follow-up measurement times and forecasts the \ac{CPHD} intensity function evolution between measurement scans \cite{Vinti_1966, Biria_2018}. The latter is carried out via the unscented transform. For the first and second test cases, 30 and 80 follow-up measurement scans are conducted, both with a measurement cadence of 15 seconds. 

The truth-model measurement set in a follow-up scan can either be empty or contain measurements from clutter objects and/or target objects. To generate a truth-model measurement set realization at each follow-up scan, for every object, we sample a measurement return using a single Bernoulli trial with the probability of success given by Eq. \ref{eqn:detection_probability_function}. If successful, a measurement is concatenated to the set, starting from empty, given by
\begin{equation}
    \boldsymbol{z}_k \sim \mathcal{N}(\boldsymbol{h}(\boldsymbol{x}_{k, \text{truth}}; \boldsymbol{p}), R)\mbox{,}
\end{equation}
where $\boldsymbol{x}_{k, \text{truth}}$ is the realized state for the catalog or target object at time step $k$.

The \ac{FOR} at the follow-up times is a rectangular grid that demarcates the projected \ac{CPHD} intensity in right-ascension and declination coordinates. The potential action set, $\mathcal{U}_k$, of the follow-up sensor consist of discretized right-ascension and declination pointing directions. We use a discretization interval equal to half of the \ac{FOV} of the follow-up sensor. This produces 474 and 285 potential actions at each follow-up scan for test cases 1 and 2, respectively.  

\subsection{Evaluation Metrics}

Here, we describe the methods for evaluation used in this study. To provide a baseline for comparison, we use a scanning technique in which the sensor \ac{FOV} is cycled over the entire \ac{FOR}, starting at portions of the search set with the highest intensity, and moving towards those with lower. The two approaches (baseline and proposed) are evaluated by comparing their multi-target state estimates to the truth-model state realization for each of 100 Monte-Carlo trials. We use an evaluation measure based on the R\'enyi-divergence between the true multi-target state and the \ac{CPHD} associated to the targets we are attempting to acquire. For $\alpha = 0.5$, this is given by
\begin{equation}
    \mathcal{D}_{0.5, k}(f_{\text{truth}}||f_{\text{est.}}) = N^*\ln{N^*} - 2N^*\ln\left(\sum_{i = 1}^{N^*}\sum_{j = 1}^{N_\text{mix.}} w_k^j \mathcal{N}(\boldsymbol{x}_{k, \text{truth}}^i; \boldsymbol{\mu}_{k}^j, P_k^j) \right) - \ln\left(\rho_k(N^*)\right)\mbox{,}\label{eqn:div_wrt_truth}
\end{equation}
where $N^*$ is the true number of targets. This form should not be confused with that introduced previously -- recall, for deciding sensor actions, we evaluated the expected divergence between the step-wise prior and posterior \ac{CPHD}. Its use here, however, is convenient as it succinctly encapsulates state and cardinality error with respect to the true multi-target state through the second and third terms, respectively. Indeed, relating to the former, the second term can be interpreted as the likelihood agreement used for uncertainty quantification in \cite{DeMars2013b} and \cite{Jones2019}. Additionally, we evaluate the expected value of the absolute cardinality error, which is given by 
\begin{equation}
    \sum_{n \geq 0} \rho_k(n)|n - N^*|\mbox{,}\label{eqn:card_error}
\end{equation}
where the operator $|\cdot|$ here denotes the absolute value.

\subsection{Results for Test Case \#1}

\begin{figure}[hbt!]
\centering
\includegraphics[width=0.95\textwidth]{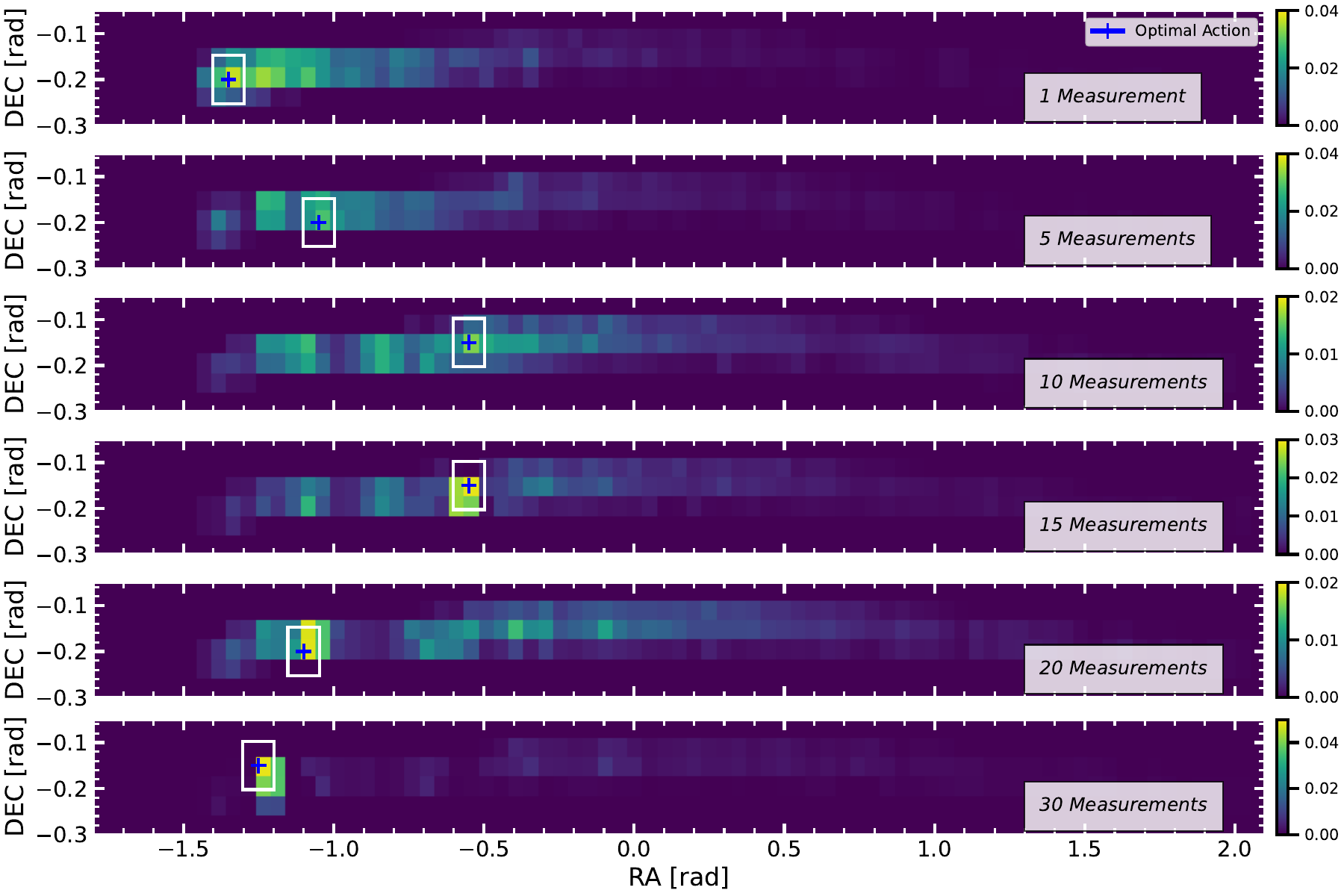}
\caption{Reward evaluated over the \ac{FOR} prior to measurement scans 1, 5, 10, 15, 20, and 30 for test case \#1. The white box denotes the \ac{FOV} and the blue cross locates the optimal action.}\label{fig:reward_hist_GTO}
\end{figure}
\begin{figure}[hbt!]
\centering
\includegraphics[width=0.95\textwidth]{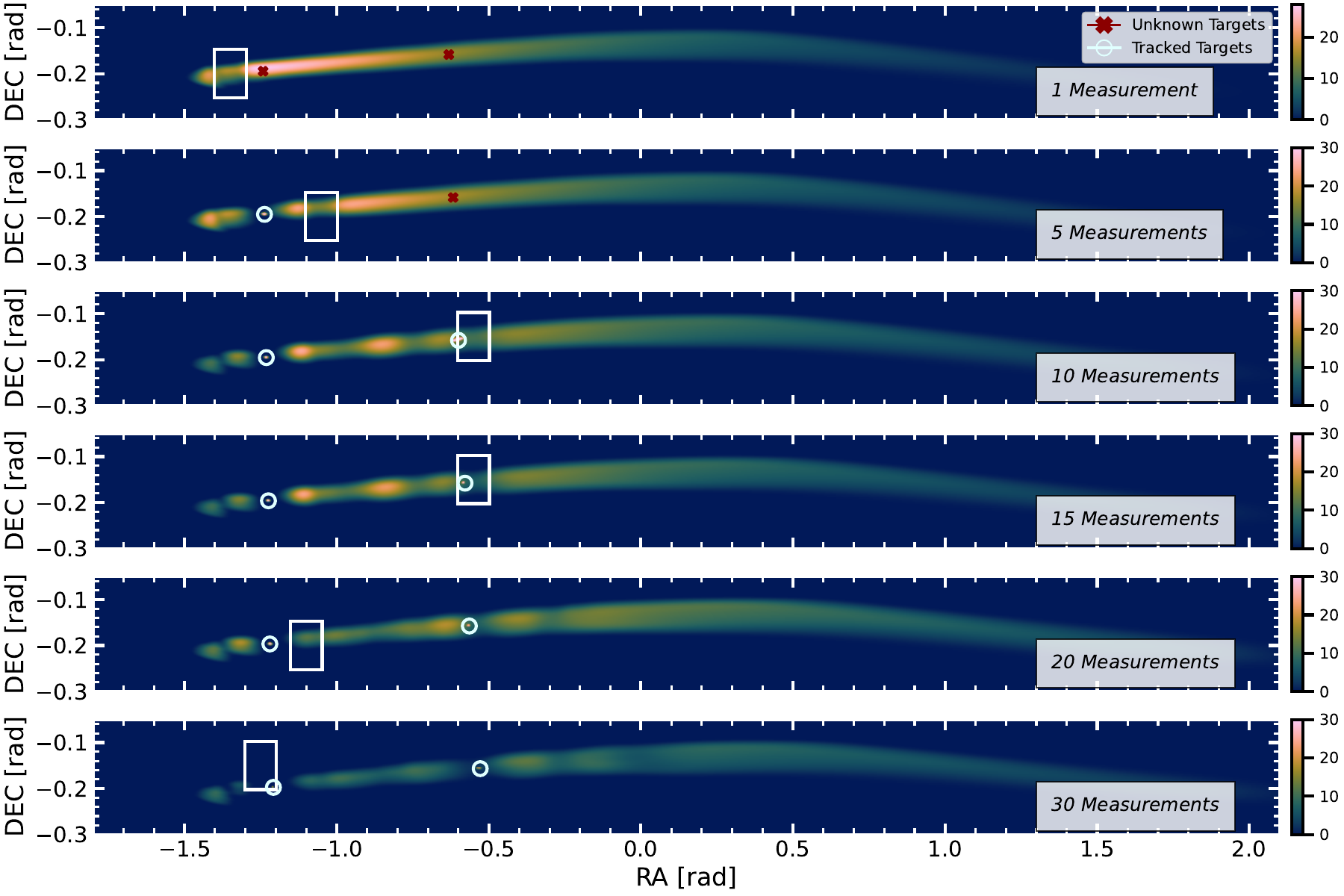}
\caption{\ac{CPHD} intensity projected into the \ac{FOR} after measurement scans 1, 5, 10, 15, 20, and 30 for test case \#1. Prior to gaining follow-up measurements, we denote targets via the red crosses. Acquired targets are circled for visibility.}\label{fig:radec_hist_GTO}
\end{figure}

We investigate the performance of our methods for test case \#1. We first consider some of the qualities of our approach for one Monte-Carlo realization. Fig. \ref{fig:reward_hist_GTO} displays the expected reward by evaluating Eq. \ref{eqn:alpha_div_CPHD_particle_approx} at each admissible sensor pointing direction over the \ac{FOR}. These reward maps are computed at time steps prior to sensor scan 1, 5, 10, 15, 20, and 30, respectively. The maximum of the reward map informs the pointing direction of the immediate next sensor scan. In each panel, the white box denotes the \ac{FOV} and the blue cross marks the optimal action.

\begin{figure}[hbt!]
\centering
\includegraphics[width=0.95\textwidth]{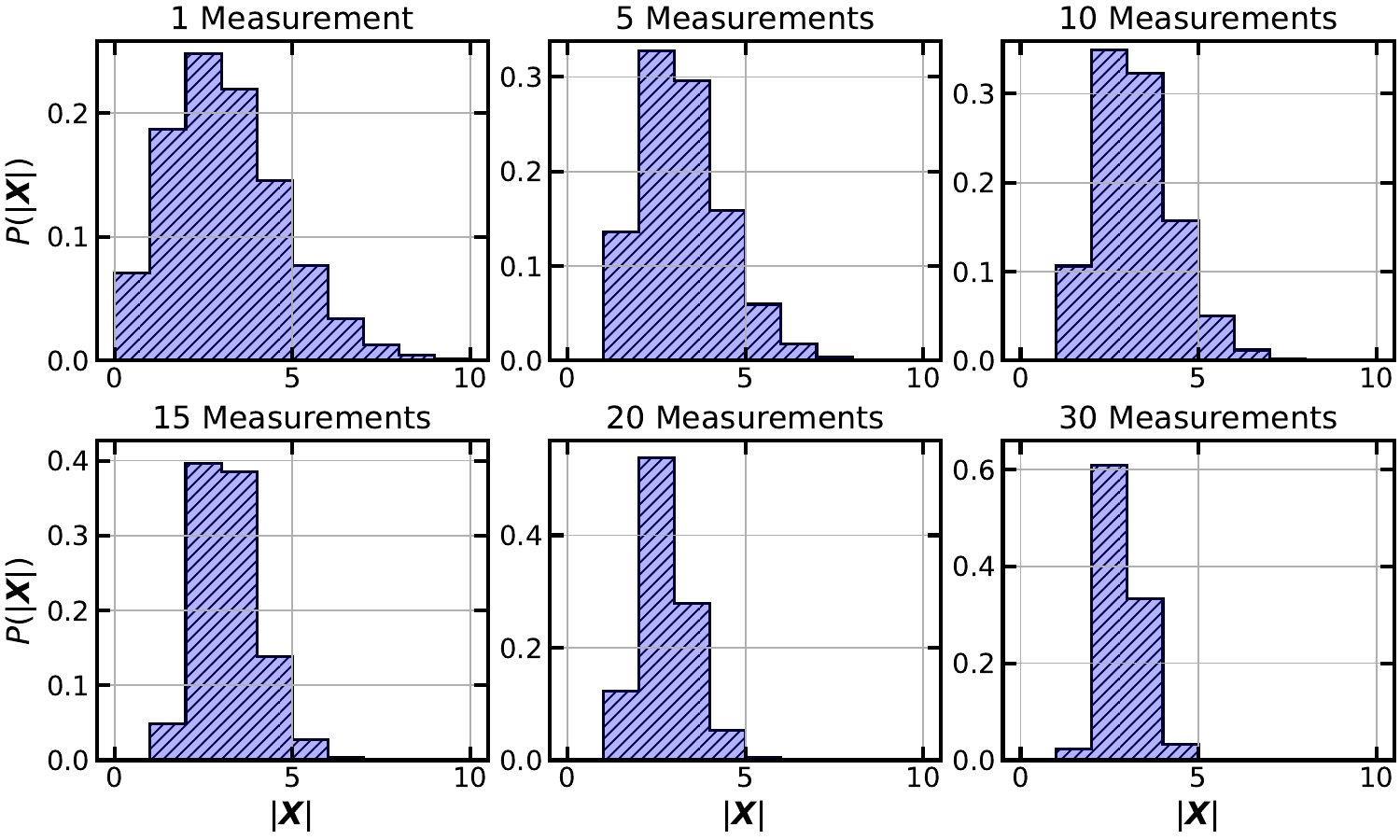}
\caption{Posterior cardinality \ac{PMF} following measurement scans 1, 5, 10, 15, 20, and 30 for test case \#1. }\label{fig:card_hist_GTO}
\end{figure}

In Fig. \ref{fig:radec_hist_GTO} we plot the impact of each action by plotting the posterior \ac{CPHD} intensity following the measurement update step. Prior to receiving a positive measurement, we mark each of the two untracked targets with a red cross and circle each following a positive measurement for visibility. We note that regions of higher intensity generally exhibit a higher expected reward. Following a positive measurement, there is a tendency of the sensor to either linger or go back to said target, as the sensing agent is incentivized through the information-driven reward to constrain the orbit of the target through tertiary measurements. In scans where the measurement set does not contain positive target measurements, the \ac{CPHD} filter acts to ``de-weight" the portion contained interior to the \ac{FOV}, commensurate with $P_D$. While the search set contained 10 catalog objects, we can robustly discriminate measurements from these clutter sources, and the filter exhibits no false tracks. In Fig. \ref{fig:card_hist_GTO} we plot the posterior cardinality \ac{PMF} for each of the aforementioned time stamps. The initial cardinality is instantiated by a Poisson distribution with a mean of 3 targets. Follow-up scans in the \ac{CPHD} recursion acts to constrain this \ac{PMF}, and the mass converges towards the true target number.  

\begin{figure}[hbt!]
\centering
\includegraphics[width=1.0\textwidth]{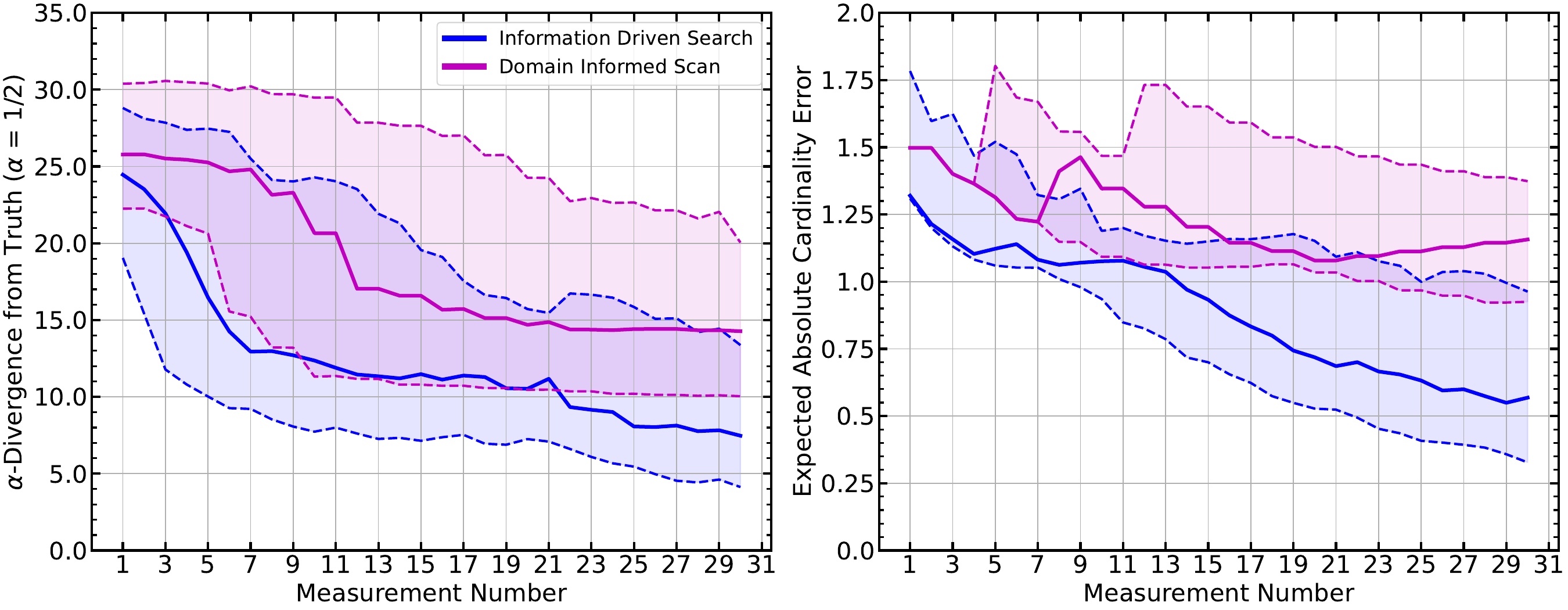}
\caption{Divergence (left) and absolute cardinality error (right) evolution for the information-driven and scanning sensor control approaches for test case \#1.}\label{fig:div_hist_GTO}
\end{figure}

In Fig. \ref{fig:div_hist_GTO} we evaluate the performance of our information-driven tasking approach and the heuristic scanning technique over 100 Monte-Carlo trials. Each trial is generated by randomly sampling true target states and clutter catalog states over the search set as described previously. We emphasize that for each trial, the number of targets and clutter objects is consistent. On the left-hand side, we evaluate Eq. \ref{eqn:div_wrt_truth} following each sequential scan and plot the median as well as the 25\%/75\% quantiles of the 100 trials. This is repeated on the right-hand side, but for the absolute cardinality error given by Eq. \ref{eqn:card_error}. For both of these metrics, the information-driven tasking strategy outperforms the simple heuristic scanning technique in both initial convergence and asymptotically. One can see that after about 17 scans, the median of the divergence tends to plateau for the scanning approach, whereas the information-driven tasking agent continues to converge towards the true multi-target state. This behavior is attributed to the tendency of the information-driven agent to follow-up with previous detected targets, whereas the former carries along its naive search. 



\subsection{Results for Test Case \#2}

\begin{figure}[hbt!]
\centering
\includegraphics[width=0.95\textwidth]{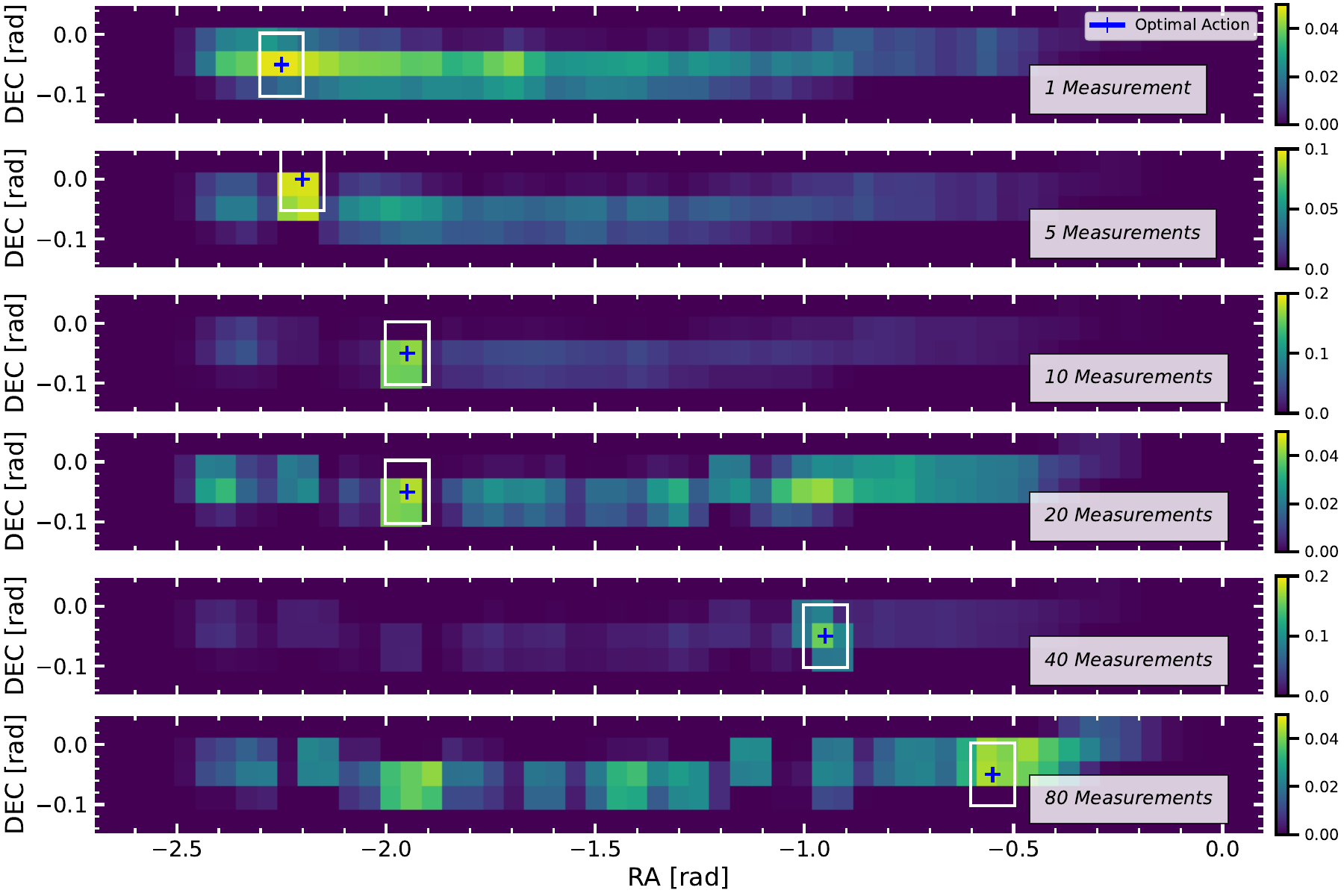}
\caption{Reward evaluated over the \ac{FOR} prior to measurement scans 1, 5, 10, 20, 40, and 80 for test case \#2. The white box denotes the \ac{FOV} and the blue cross locates the optimal action.}\label{fig:reward_hist_GEO}
\end{figure}

\begin{figure}[hbt!]
\centering
\includegraphics[width=0.95\textwidth]{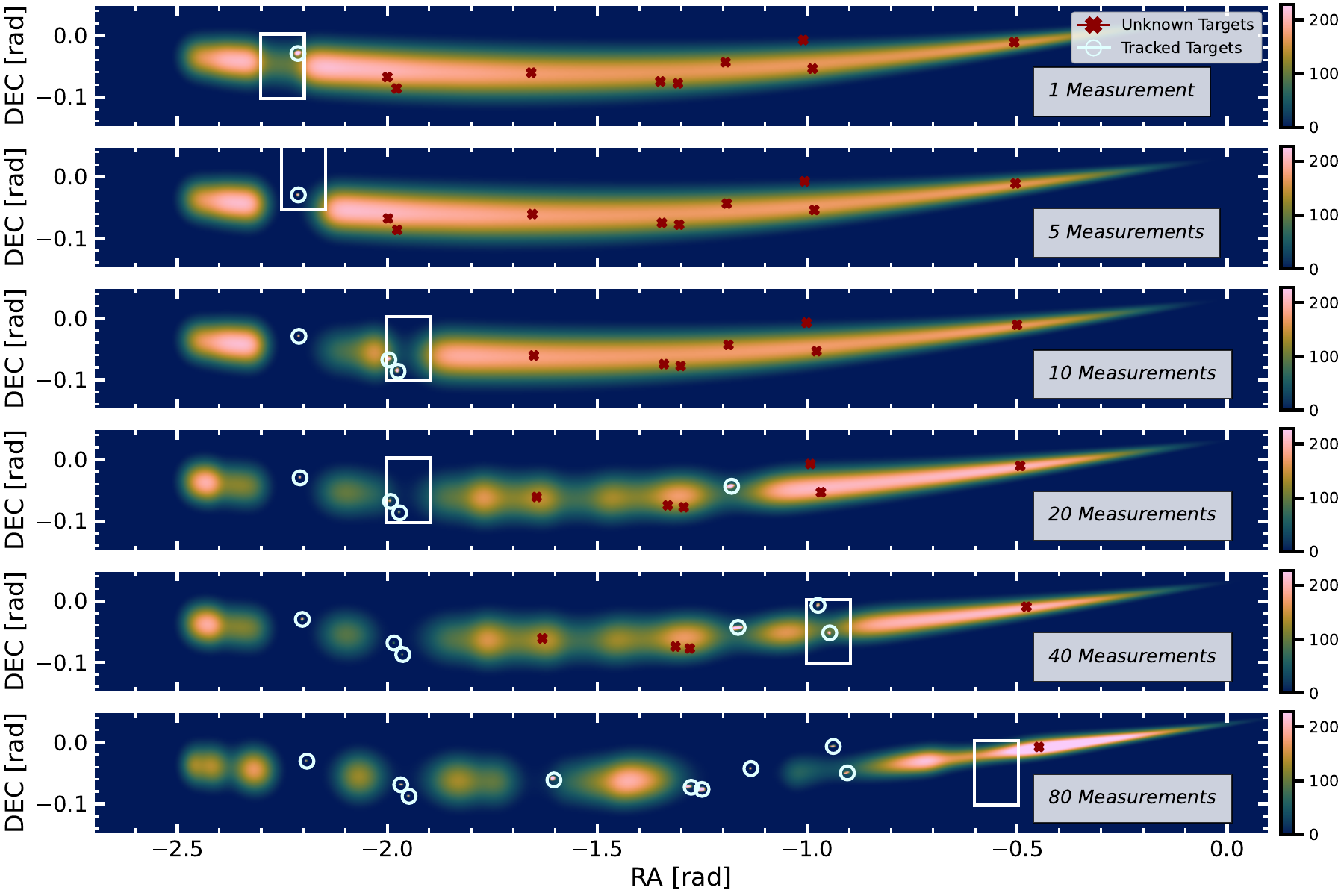}
\caption{\ac{CPHD} intensity projected into the \ac{FOR} after measurement scans 1, 5, 10, 20, 40, and 80 for test case \#2. Prior to gaining follow-up measurements, we denote targets via the red crosses. Acquired targets are circled for visibility.}\label{fig:radec_hist_GEO}
\end{figure}

\begin{figure}[hbt!]
\centering
\includegraphics[width=0.95\textwidth]{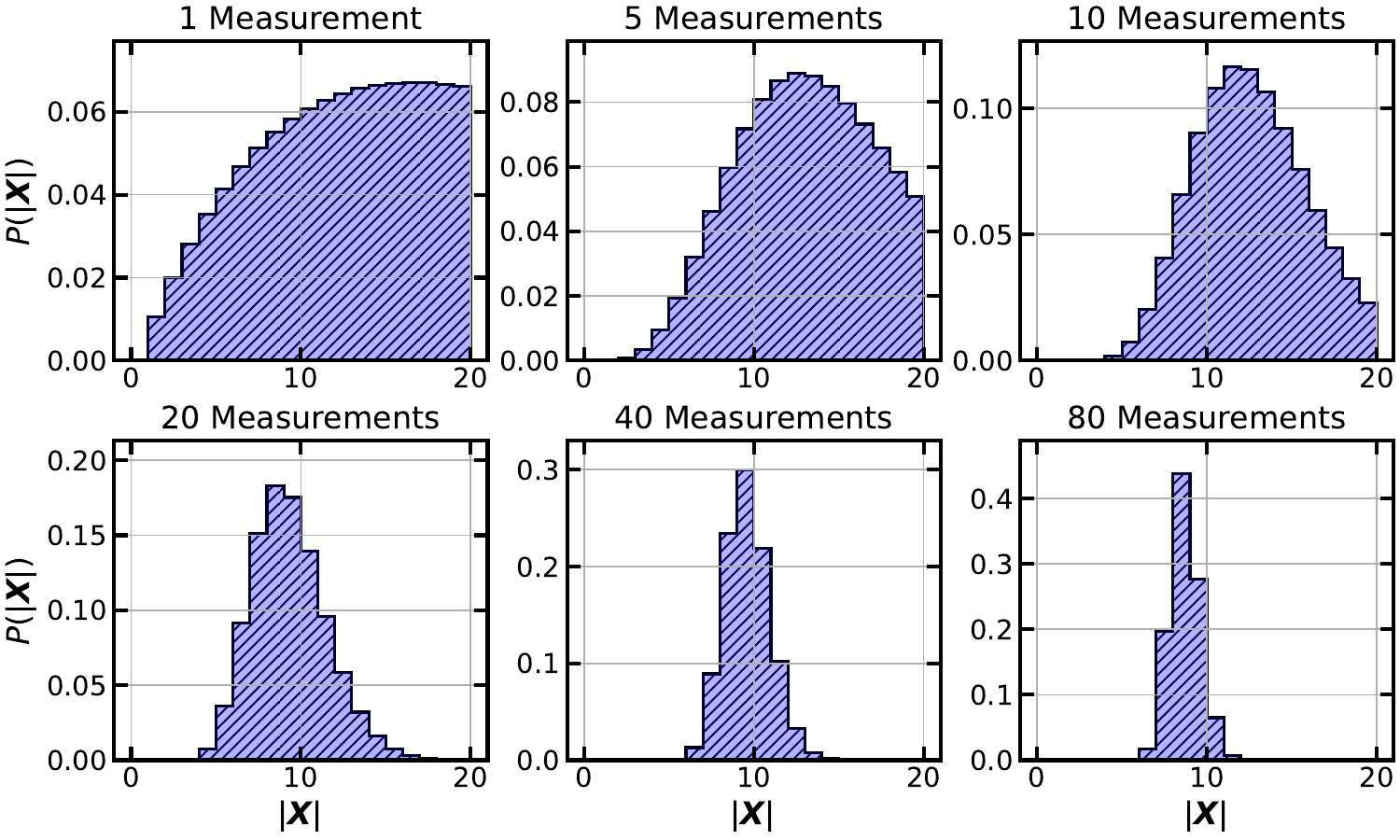}
\caption{Posterior cardinality \ac{PMF} following measurement scans 1, 5, 10, 20, 40, and 80 for test case \#2. }\label{fig:card_hist_GEO}
\end{figure}

This section evaluates the performance of the proposed techniques for the second test case. We consider this case more challenging as it includes more catalog objects and untracked targets; 15 and 10, respectively. Moreover, we assume no prior knowledge of the true target number by instantiating the \ac{CPHD} cardinality \ac{PMF} with a uniform mass between 0 and 19 targets. For these reasons, we increase the number of measurement scans to 80. Similar to before, in Fig. \ref{fig:reward_hist_GEO}, we plot the expected reward prior to the measurement scans 1, 5, 10, 20, 40 and 80. Comparing with Fig. \ref{fig:radec_hist_GEO}, we can see in scan \#1 regions with higher intensity generally correlate with higher expected reward. Upon target localization, there is a marked increase in the expected reward at said location, which, as noted previously, corresponds to the expected information gain produced by further constraining the orbit of the target(s). Inspecting the tail-end of the search set in the bottom panel of Fig. \ref{fig:radec_hist_GEO}, we note the increase in intensity relative to prior scans. This phenomenon is often referred to as the ``spooky effect" in the multi-target tracking community \cite{Vo2012}, and while it is generally seen as a fallacy of the \ac{CPHD} filter, is advantageous for our application as it incentives the sensing agent to visit those portions not yet scanned. Following 80 measurement scans, the sensor has successfully localized 9 of the 10 true targets with no false tracks. In Fig. \ref{fig:card_hist_GEO} we plot the cardinality \ac{PMF} evolution for this test case for each of the aforementioned time stamps. While this is initialized with a diffuse prior, the \ac{CPHD} converges close to the true cardinality after 80 scans.

\begin{figure}[hbt!]
\centering
\includegraphics[width=1.0\textwidth]{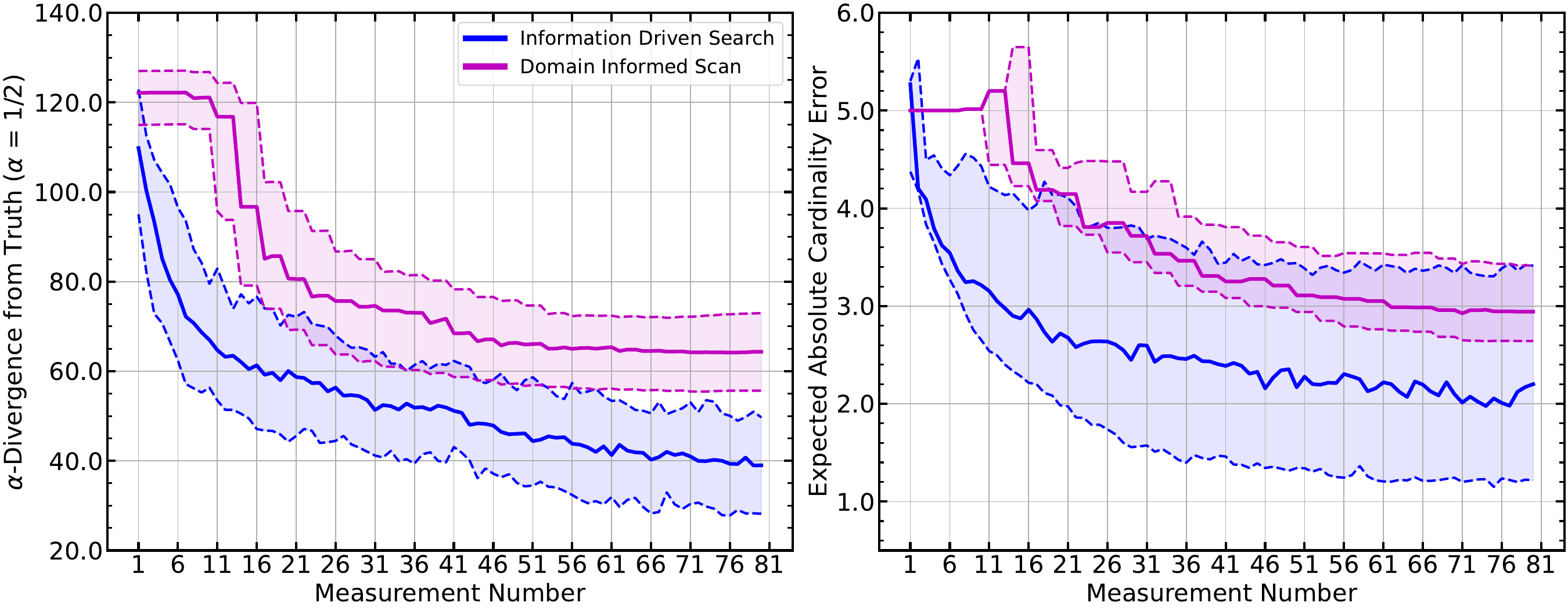}
\caption{\textbf{Divergence (left) and absolute cardinality error (right) evolution for the information-driven and scanning sensor control approaches for test case \#2.}}\label{fig:div_hist_GEO}
\end{figure}

Like before, Fig. \ref{fig:div_hist_GEO} evaluates the divergence and absolute cardinality error over the 100 Monte-Carlo trials, but for test case \#2. We observe again better initial and asymptotic performance in these metrics for our information-driven tasking over a naive search strategy. While the naive search plateaus after about 50 measurements, our proposed tasking strategy continues to converge in localization accuracy, seen in the continued decrease in divergence. It is interesting, however, to note the 25\%/75\% quantile spread in cardinality error for the information driven approach (upper and lower blue dashed lines). This result can be attributed to a lower variance in the posterior \ac{CPHD} cardinality \ac{PMF} produced by this tasking strategy. In cases where the mode does not directly coincide with the true target number, this causes a higher cardinality error with Eq. \ref{eqn:card_error}, and on the other hand, a much lower error when the \ac{PMF} mode does directly coincide. Over the 100 Monte-Carlo trials, we see the effect of this in this spread.

%
\section{Conclusions}

This work introduces a closed loop multi-target filtering and sensor control approach for gaining custody of high-priority untracked space objects. A search set is instantiated with a novel measurement using the \ac{AR} method and a \ac{GMM} parameterization of the multi-target \ac{CPHD} intensity function. From this, the set is forecast to a future time for hand-off to a follow-up optical sensor. In follow-up scans, the \ac{CPHD} description of the multi-target state allows us to formally consider multiple targets within the search set, false-positive and missed detections, and information contained in empty measurements. Moreover, the \ac{CPHD} description is compatible with our clutter model that accounts for the false-positive measurements arising from known catalog objects. To account for the discontinuity in the detection probability function between portions of the set inside and outside the \ac{FOV}, we introduce a \ac{GMM} splitting procedure amenable for nonlinear maps between the state space and \ac{FOR}. Optimal sensor steering is formulated as a \ac{POMDP} in which the latest information state is the \ac{CPHD} associated to the untracked targets, and the sensor actions encode the angular pointing directions of the optical telescope. The reward function of the tasking agent is taken to be an information theoretic function that measures the expected information gained in a potential follow-up sensor action. We test our methods with several synthetic case studies of relevance to the \ac{SDA} community and find that the proposed approach can significantly outperform conventional naive tasking strategies in several methods for quickly and accuracy gaining custody of novel space objects. 

While the methods presented here are adapted for optical telescopes, they can more generally be applied to sensing resources of different modalities. For instance, within the context of rapid acquisition, the two-short-arc time latency associated with strictly optical measurements can be circumvented by adding range information. Future work intends to generalize the proposed techniques developed here for these non-traditional sensor configurations.


\bmhead{Acknowledgements}

The authors thank Michael Steinbock and Ryan Coder for their helpful guidance and suggestions. This material is based on research sponsored by Air Force Research Laboratory (AFRL) under agreement number FA9453-21-2-0064. The U.S. Government is authorized to reproduce and distribute reprints for Governmental purposes notwithstanding any copyright notation thereon.  

The views and conclusions contained herein are those of the authors and should not be interpreted as necessarily representing the official policies or endorsements, either expressed or implied, of Air Force Research Laboratory (AFRL) and or the U.S. Government.  AFRL Public Release Case Number:  AFRL-2024-4849.

\section*{Conflicts of Interest}

On behalf of all authors, the corresponding author states that there is no conflict of interest.

\backmatter

\bibliography{references}


\begin{thebibliography}{37}
\ifx \bisbn   \undefined \def \bisbn  #1{ISBN #1}\fi
\ifx \binits  \undefined \def \binits#1{#1}\fi
\ifx \bauthor  \undefined \def \bauthor#1{#1}\fi
\ifx \batitle  \undefined \def \batitle#1{#1}\fi
\ifx \bjtitle  \undefined \def \bjtitle#1{#1}\fi
\ifx \bvolume  \undefined \def \bvolume#1{\textbf{#1}}\fi
\ifx \byear  \undefined \def \byear#1{#1}\fi
\ifx \bissue  \undefined \def \bissue#1{#1}\fi
\ifx \bfpage  \undefined \def \bfpage#1{#1}\fi
\ifx \blpage  \undefined \def \blpage #1{#1}\fi
\ifx \burl  \undefined \def \burl#1{\textsf{#1}}\fi
\ifx \doiurl  \undefined \def \doiurl#1{\url{https://doi.org/#1}}\fi
\ifx \betal  \undefined \def \betal{\textit{et al.}}\fi
\ifx \binstitute  \undefined \def \binstitute#1{#1}\fi
\ifx \binstitutionaled  \undefined \def \binstitutionaled#1{#1}\fi
\ifx \bctitle  \undefined \def \bctitle#1{#1}\fi
\ifx \beditor  \undefined \def \beditor#1{#1}\fi
\ifx \bpublisher  \undefined \def \bpublisher#1{#1}\fi
\ifx \bbtitle  \undefined \def \bbtitle#1{#1}\fi
\ifx \bedition  \undefined \def \bedition#1{#1}\fi
\ifx \bseriesno  \undefined \def \bseriesno#1{#1}\fi
\ifx \blocation  \undefined \def \blocation#1{#1}\fi
\ifx \bsertitle  \undefined \def \bsertitle#1{#1}\fi
\ifx \bsnm \undefined \def \bsnm#1{#1}\fi
\ifx \bsuffix \undefined \def \bsuffix#1{#1}\fi
\ifx \bparticle \undefined \def \bparticle#1{#1}\fi
\ifx \barticle \undefined \def \barticle#1{#1}\fi
\bibcommenthead
\ifx \bconfdate \undefined \def \bconfdate #1{#1}\fi
\ifx \botherref \undefined \def \botherref #1{#1}\fi
\ifx \url \undefined \def \url#1{\textsf{#1}}\fi
\ifx \bchapter \undefined \def \bchapter#1{#1}\fi
\ifx \bbook \undefined \def \bbook#1{#1}\fi
\ifx \bcomment \undefined \def \bcomment#1{#1}\fi
\ifx \oauthor \undefined \def \oauthor#1{#1}\fi
\ifx \citeauthoryear \undefined \def \citeauthoryear#1{#1}\fi
\ifx \endbibitem  \undefined \def \endbibitem {}\fi
\ifx \bconflocation  \undefined \def \bconflocation#1{#1}\fi
\ifx \arxivurl  \undefined \def \arxivurl#1{\textsf{#1}}\fi
\csname PreBibitemsHook\endcsname

\bibitem[\protect\citeauthoryear{Milani et~al.}{2004}]{Milani2004}
\begin{barticle}
\bauthor{\bsnm{Milani}, \binits{A.}},
\bauthor{\bsnm{Gronchi}, \binits{G.F.}},
\bauthor{\bsnm{Vitturi}, \binits{M.D.M.}},
\bauthor{\bsnm{Kne{\v{z}}evi{\'c}}, \binits{Z.}}:
\batitle{Orbit determination with very short arcs. i admissible regions}.
\bjtitle{Celestial Mechanics and Dynamical Astronomy}
\bvolume{90},
\bfpage{57}--\blpage{85}
(\byear{2004})
\end{barticle}
\endbibitem

\bibitem[\protect\citeauthoryear{Fujimoto and Scheeres}{2013}]{Fujimoto2013}
\begin{barticle}
\bauthor{\bsnm{Fujimoto}, \binits{K.}},
\bauthor{\bsnm{Scheeres}, \binits{D.J.}}:
\batitle{Applications of the admissible region to space-based observations}.
\bjtitle{Advances in Space Research}
\bvolume{52}(\bissue{4}),
\bfpage{696}--\blpage{704}
(\byear{2013})
\doiurl{10.1016/j.asr.2013.04.020}
\end{barticle}
\endbibitem

\bibitem[\protect\citeauthoryear{DeMars and Jah}{2013}]{DeMars2013}
\begin{barticle}
\bauthor{\bsnm{DeMars}, \binits{K.J.}},
\bauthor{\bsnm{Jah}, \binits{M.K.}}:
\batitle{Probabilistic initial orbit determination using {G}aussian mixture models}.
\bjtitle{Journal of Guidance, Control, and Dynamics}
\bvolume{36}(\bissue{5}),
\bfpage{1324}--\blpage{1335}
(\byear{2013})
\doiurl{10.2514/1.59844}
{\href{https://arxiv.org/abs/https://doi.org/10.2514/1.59844}{{https://doi.org/10.2514/1.59844}}}
\end{barticle}
\endbibitem

\bibitem[\protect\citeauthoryear{Mahler}{2014}]{Mahler2014}
\begin{bbook}
\beditor{\bsnm{Mahler}, \binits{R.P.S.}} (ed.):
\bbtitle{Advances in Statistical Multisource-Multitarget Information Fusion},
pp. \bfpage{827}--\blpage{939}.
\bpublisher{Artech House},
\blocation{Norwood, MA}
(\byear{2014})
\end{bbook}
\endbibitem

\bibitem[\protect\citeauthoryear{Mahler}{2003}]{Mahler2003}
\begin{barticle}
\bauthor{\bsnm{Mahler}, \binits{R.P.S.}}:
\batitle{Multitarget {B}ayes filtering via first-order multitarget moments}.
\bjtitle{IEEE Transactions on Aerospace and Electronic Systems}
\bvolume{39}(\bissue{4}),
\bfpage{1152}--\blpage{1178}
(\byear{2003})
\doiurl{10.1109/TAES.2003.1261119}
\end{barticle}
\endbibitem

\bibitem[\protect\citeauthoryear{{Murphy} and {Holzinger}}{2017}]{Murphy2017}
\begin{bchapter}
\bauthor{\bsnm{{Murphy}}, \binits{T.S.}},
\bauthor{\bsnm{{Holzinger}}, \binits{M.J.}}:
\bctitle{Generalized minimum-time follow-up approaches applied to tasking electro-optical sensor tasking}.
In: \beditor{\bsnm{{Ryan}}, \binits{S.}} (ed.)
\bbtitle{Advanced Maui Optical and Space Surveillance (AMOS) Technologies Conference},
p. \bfpage{54}
(\byear{2017})
\end{bchapter}
\endbibitem

\bibitem[\protect\citeauthoryear{{Fedeler} et~al.}{2022}]{Fedeler2022}
\begin{barticle}
\bauthor{\bsnm{{Fedeler}}, \binits{S.J.}},
\bauthor{\bsnm{{Holzinger}}, \binits{M.J.}},
\bauthor{\bsnm{{Whitacre}}, \binits{W.W.}}:
\batitle{Tasking and estimation for minimum-time space object search and recovery}.
\bjtitle{Journal of the Astronautical Sciences}
\bvolume{69}(\bissue{4}),
\bfpage{1216}--\blpage{1249}
(\byear{2022})
\doiurl{10.1007/s40295-022-00332-0}
\end{barticle}
\endbibitem

\bibitem[\protect\citeauthoryear{LeGrand and Ferrari}{2022}]{LeGrand2022}
\begin{botherref}
\oauthor{\bsnm{LeGrand}, \binits{K.A.}},
\oauthor{\bsnm{Ferrari}, \binits{S.}}:
Split happens! imprecise and negative information in gaussian mixture random finite set filtering.
Journal of Advances in Information Fusion
\textbf{17}(2)
(2022)
\end{botherref}
\endbibitem

\bibitem[\protect\citeauthoryear{Ristic et~al.}{2011}]{Ristic2011}
\begin{barticle}
\bauthor{\bsnm{Ristic}, \binits{B.}},
\bauthor{\bsnm{Vo}, \binits{B.-N.}},
\bauthor{\bsnm{Clark}, \binits{D.}}:
\batitle{A note on the reward function for {PHD} filters with sensor control}.
\bjtitle{IEEE Transactions on Aerospace and Electronic Systems}
\bvolume{47}(\bissue{2}),
\bfpage{1521}--\blpage{1529}
(\byear{2011})
\doiurl{10.1109/TAES.2011.5751278}
\end{barticle}
\endbibitem

\bibitem[\protect\citeauthoryear{Cai and Ferrari}{2009}]{Cai2009}
\begin{barticle}
\bauthor{\bsnm{Cai}, \binits{C.}},
\bauthor{\bsnm{Ferrari}, \binits{S.}}:
\batitle{Information-driven sensor path planning by approximate cell decomposition}.
\bjtitle{IEEE Transactions on Systems, Man, and Cybernetics, Part B (Cybernetics)}
\bvolume{39}(\bissue{3}),
\bfpage{672}--\blpage{689}
(\byear{2009})
\doiurl{10.1109/TSMCB.2008.2008561}
\end{barticle}
\endbibitem

\bibitem[\protect\citeauthoryear{Beard et~al.}{2017}]{Beard2015}
\begin{barticle}
\bauthor{\bsnm{Beard}, \binits{M.}},
\bauthor{\bsnm{Vo}, \binits{B.-T.}},
\bauthor{\bsnm{Vo}, \binits{B.-N.}},
\bauthor{\bsnm{Arulampalam}, \binits{S.}}:
\batitle{Void probabilities and {C}auchy–{S}chwarz divergence for generalized labeled multi-bernoulli models}.
\bjtitle{IEEE Transactions on Signal Processing}
\bvolume{65}(\bissue{19}),
\bfpage{5047}--\blpage{5061}
(\byear{2017})
\doiurl{10.1109/TSP.2017.2723355}
\end{barticle}
\endbibitem

\bibitem[\protect\citeauthoryear{LeGrand et~al.}{2021}]{LeGrand2021}
\begin{bchapter}
\bauthor{\bsnm{LeGrand}, \binits{K.A.}},
\bauthor{\bsnm{Zhu}, \binits{P.}},
\bauthor{\bsnm{Ferrari}, \binits{S.}}:
\bctitle{A random finite set sensor control approach for vision-based multi-object search-while-tracking}.
In: \bbtitle{2021 IEEE 24th International Conference on Information Fusion (FUSION)},
pp. \bfpage{1}--\blpage{8}
(\byear{2021}).
\doiurl{10.23919/FUSION49465.2021.9626898}
\end{bchapter}
\endbibitem

\bibitem[\protect\citeauthoryear{Worthy and Holzinger}{2015}]{Worthy2015}
\begin{barticle}
\bauthor{\bsnm{Worthy}, \binits{J.L.}},
\bauthor{\bsnm{Holzinger}, \binits{M.J.}}:
\batitle{Incorporating uncertainty in admissible regions for uncorrelated detections}.
\bjtitle{Journal of Guidance, Control, and Dynamics}
\bvolume{38}(\bissue{9}),
\bfpage{1673}--\blpage{1689}
(\byear{2015})
\doiurl{10.2514/1.G000890}
{\href{https://arxiv.org/abs/https://doi.org/10.2514/1.G000890}{{https://doi.org/10.2514/1.G000890}}}
\end{barticle}
\endbibitem

\bibitem[\protect\citeauthoryear{Gehly et~al.}{2018}]{Gehly2018}
\begin{barticle}
\bauthor{\bsnm{Gehly}, \binits{S.}},
\bauthor{\bsnm{Jones}, \binits{B.A.}},
\bauthor{\bsnm{Axelrad}, \binits{P.}}:
\batitle{Search-detect-track sensor allocation for geosynchronous space objects}.
\bjtitle{IEEE Transactions on Aerospace and Electronic Systems}
\bvolume{54}(\bissue{6}),
\bfpage{2788}--\blpage{2808}
(\byear{2018})
\doiurl{10.1109/TAES.2018.2830578}
\end{barticle}
\endbibitem

\bibitem[\protect\citeauthoryear{Wan and Van Der~Merwe}{2000}]{Wan2000}
\begin{bchapter}
\bauthor{\bsnm{Wan}, \binits{E.A.}},
\bauthor{\bsnm{Van Der~Merwe}, \binits{R.}}:
\bctitle{The unscented {K}alman filter for nonlinear estimation}.
In: \bbtitle{Proceedings of the IEEE 2000 Adaptive Systems for Signal Processing, Communications, and Control Symposium (Cat. No.00EX373)},
pp. \bfpage{153}--\blpage{158}
(\byear{2000}).
\doiurl{10.1109/ASSPCC.2000.882463}
\end{bchapter}
\endbibitem

\bibitem[\protect\citeauthoryear{Mahler}{2007}]{Mahler2007}
\begin{barticle}
\bauthor{\bsnm{Mahler}, \binits{R.}}:
\batitle{{PHD} filters of higher order in target number}.
\bjtitle{IEEE Transactions on Aerospace and Electronic Systems}
\bvolume{43}(\bissue{4}),
\bfpage{1523}--\blpage{1543}
(\byear{2007})
\doiurl{10.1109/TAES.2007.4441756}
\end{barticle}
\endbibitem

\bibitem[\protect\citeauthoryear{Vo et~al.}{2007}]{Vo2007}
\begin{barticle}
\bauthor{\bsnm{Vo}, \binits{B.-T.}},
\bauthor{\bsnm{Vo}, \binits{B.-N.}},
\bauthor{\bsnm{Cantoni}, \binits{A.}}:
\batitle{Analytic implementations of the cardinalized probability hypothesis density filter}.
\bjtitle{IEEE Transactions on Signal Processing}
\bvolume{55}(\bissue{7}),
\bfpage{3553}--\blpage{3567}
(\byear{2007})
\doiurl{10.1109/TSP.2007.894241}
\end{barticle}
\endbibitem

\bibitem[\protect\citeauthoryear{Ristic et~al.}{2012}]{Ristic2012}
\begin{barticle}
\bauthor{\bsnm{Ristic}, \binits{B.}},
\bauthor{\bsnm{Clark}, \binits{D.}},
\bauthor{\bsnm{Vo}, \binits{B.-N.}},
\bauthor{\bsnm{Vo}, \binits{B.-T.}}:
\batitle{Adaptive target birth intensity for {PHD} and {CPHD} filters}.
\bjtitle{IEEE Transactions on Aerospace and Electronic Systems}
\bvolume{48}(\bissue{2}),
\bfpage{1656}--\blpage{1668}
(\byear{2012})
\doiurl{10.1109/TAES.2012.6178085}
\end{barticle}
\endbibitem

\bibitem[\protect\citeauthoryear{Van~Assche}{1996}]{Assche1996}
\begin{barticle}
\bauthor{\bsnm{Van~Assche}, \binits{W.}}:
\batitle{Polynomials and polynomial inequalities (p. borwein and t. erdélyi)}.
\bjtitle{SIAM Review}
\bvolume{38}(\bissue{4}),
\bfpage{705}--\blpage{706}
(\byear{1996})
\doiurl{10.1137/1038150}
\end{barticle}
\endbibitem

\bibitem[\protect\citeauthoryear{DeMars et~al.}{2013}]{DeMars2013b}
\begin{barticle}
\bauthor{\bsnm{DeMars}, \binits{K.J.}},
\bauthor{\bsnm{Bishop}, \binits{R.H.}},
\bauthor{\bsnm{Jah}, \binits{M.K.}}:
\batitle{Entropy-based approach for uncertainty propagation of nonlinear dynamical systems}.
\bjtitle{Journal of Guidance, Control, and Dynamics}
\bvolume{36}(\bissue{4}),
\bfpage{1047}--\blpage{1057}
(\byear{2013})
\doiurl{10.2514/1.58987}
{\href{https://arxiv.org/abs/https://doi.org/10.2514/1.58987}{{https://doi.org/10.2514/1.58987}}}
\end{barticle}
\endbibitem

\bibitem[\protect\citeauthoryear{{Wolf} and {Jones}}{2023}]{Wolf2023}
\begin{bchapter}
\bauthor{\bsnm{{Wolf}}, \binits{T.}},
\bauthor{\bsnm{{Jones}}, \binits{B.}}:
\bctitle{Intelligent sensor tasking for minimum-time space object acquisition}.
In: \beditor{\bsnm{{Ryan}}, \binits{S.}} (ed.)
\bbtitle{Proceedings of the Advanced Maui Optical and Space Surveillance (AMOS) Technologies Conference},
p. \bfpage{200}
(\byear{2023})
\end{bchapter}
\endbibitem

\bibitem[\protect\citeauthoryear{Sunberg et~al.}{2016}]{Sunberg2016}
\begin{barticle}
\bauthor{\bsnm{Sunberg}, \binits{Z.}},
\bauthor{\bsnm{Chakravorty}, \binits{S.}},
\bauthor{\bsnm{Erwin}, \binits{R.S.}}:
\batitle{Information space receding horizon control for multisensor tasking problems}.
\bjtitle{IEEE Transactions on Cybernetics}
\bvolume{46}(\bissue{6}),
\bfpage{1325}--\blpage{1336}
(\byear{2016})
\doiurl{10.1109/TCYB.2015.2445744}
\end{barticle}
\endbibitem

\bibitem[\protect\citeauthoryear{Adurthi et~al.}{2020}]{Adurthi2020}
\begin{barticle}
\bauthor{\bsnm{Adurthi}, \binits{N.}},
\bauthor{\bsnm{Singla}, \binits{P.}},
\bauthor{\bsnm{Majji}, \binits{M.}}:
\batitle{Mutual information based sensor tasking with applications to space situational awareness}.
\bjtitle{Journal of Guidance, Control, and Dynamics}
\bvolume{43}(\bissue{4}),
\bfpage{767}--\blpage{789}
(\byear{2020})
\doiurl{10.2514/1.G004399}
{\href{https://arxiv.org/abs/https://doi.org/10.2514/1.G004399}{{https://doi.org/10.2514/1.G004399}}}
\end{barticle}
\endbibitem

\bibitem[\protect\citeauthoryear{van Erven and Harremos}{2014}]{vanErven2014}
\begin{barticle}
\bauthor{\bsnm{Erven}, \binits{T.}},
\bauthor{\bsnm{Harremos}, \binits{P.}}:
\batitle{{R}ényi divergence and {K}ullback-{L}eibler divergence}.
\bjtitle{IEEE Transactions on Information Theory}
\bvolume{60}(\bissue{7}),
\bfpage{3797}--\blpage{3820}
(\byear{2014})
\doiurl{10.1109/TIT.2014.2320500}
\end{barticle}
\endbibitem

\bibitem[\protect\citeauthoryear{Hero et~al.}{2008}]{Hero2008}
\begin{bbook}
\bauthor{\bsnm{Hero}, \binits{A.O.}},
\bauthor{\bsnm{Kreucher}, \binits{C.M.}},
\bauthor{\bsnm{Blatt}, \binits{D.}}:
In: \beditor{\bsnm{Hero}, \binits{A.O.}},
\beditor{\bsnm{Casta{\~{n}}{\'o}n}, \binits{D.A.}},
\beditor{\bsnm{Cochran}, \binits{D.}},
\beditor{\bsnm{Kastella}, \binits{K.}} (eds.)
\bbtitle{Information Theoretic Approaches to Sensor Management},
pp. \bfpage{33}--\blpage{57}.
\bpublisher{Springer},
\blocation{Boston, MA}
(\byear{2008}).
\doiurl{10.1007/978-0-387-49819-5_3} .
\burl{https://doi.org/10.1007/978-0-387-49819-5_3}
\end{bbook}
\endbibitem

\bibitem[\protect\citeauthoryear{Hershey and Olsen}{2007}]{Hershey2007}
\begin{bchapter}
\bauthor{\bsnm{Hershey}, \binits{J.R.}},
\bauthor{\bsnm{Olsen}, \binits{P.A.}}:
\bctitle{Approximating the {K}ullback {L}eibler divergence between {G}aussian mixture models}.
In: \bbtitle{2007 IEEE International Conference on Acoustics, Speech and Signal Processing - ICASSP '07},
vol. \bseriesno{4},
pp. \bfpage{317}--\blpage{320}
(\byear{2007}).
\doiurl{10.1109/ICASSP.2007.366913}
\end{bchapter}
\endbibitem

\bibitem[\protect\citeauthoryear{Huber et~al.}{2008}]{Huber2008}
\begin{bchapter}
\bauthor{\bsnm{Huber}, \binits{M.F.}},
\bauthor{\bsnm{Bailey}, \binits{T.}},
\bauthor{\bsnm{Durrant-Whyte}, \binits{H.}},
\bauthor{\bsnm{Hanebeck}, \binits{U.D.}}:
\bctitle{On entropy approximation for {G}aussian mixture random vectors}.
In: \bbtitle{2008 IEEE International Conference on Multisensor Fusion and Integration for Intelligent Systems},
pp. \bfpage{181}--\blpage{188}
(\byear{2008}).
\doiurl{10.1109/MFI.2008.4648062}
\end{bchapter}
\endbibitem

\bibitem[\protect\citeauthoryear{Skoglar et~al.}{2009}]{Skoglar2009}
\begin{bchapter}
\bauthor{\bsnm{Skoglar}, \binits{P.}},
\bauthor{\bsnm{Orguner}, \binits{U.}},
\bauthor{\bsnm{Gustafsson}, \binits{F.}}:
\bctitle{On information measures based on particle mixture for optimal bearings-only tracking}.
In: \bbtitle{2009 IEEE Aerospace Conference},
pp. \bfpage{1}--\blpage{14}
(\byear{2009}).
\doiurl{10.1109/AERO.2009.4839487}
\end{bchapter}
\endbibitem

\bibitem[\protect\citeauthoryear{Leonenko et~al.}{}]{Leoneko2008}
\begin{botherref}
\oauthor{\bsnm{Leonenko}, \binits{N.}},
\oauthor{\bsnm{Pronzato}, \binits{L.}},
\oauthor{\bsnm{Savani}, \binits{V.}}:
{A class of Rényi information estimators for multidimensional densities}.
The Annals of Statistics
\textbf{36}(5),
2153--2182
\doiurl{10.1214/07-AOS539}
\end{botherref}
\endbibitem

\bibitem[\protect\citeauthoryear{Ajgl and Šimandl}{2011}]{Jiri2011}
\begin{barticle}
\bauthor{\bsnm{Ajgl}, \binits{J.}},
\bauthor{\bsnm{Šimandl}, \binits{M.}}:
\batitle{Differential entropy estimation by particles}.
\bjtitle{IFAC Proceedings Volumes}
\bvolume{44}(\bissue{1}),
\bfpage{11991}--\blpage{11996}
(\byear{2011})
\doiurl{10.3182/20110828-6-IT-1002.01404} .
\bcomment{18th IFAC World Congress}
\end{barticle}
\endbibitem

\bibitem[\protect\citeauthoryear{{Choi} et~al.}{2017}]{Choi2017}
\begin{barticle}
\bauthor{\bsnm{{Choi}}, \binits{E.-J.}},
\bauthor{\bsnm{{Cho}}, \binits{S.}},
\bauthor{\bsnm{{Jo}}, \binits{J.H.}},
\bauthor{\bsnm{{Park}}, \binits{J.-H.}},
\bauthor{\bsnm{{Chung}}, \binits{T.}},
\bauthor{\bsnm{{Park}}, \binits{J.}},
\bauthor{\bsnm{{Jeon}}, \binits{H.}},
\bauthor{\bsnm{{Yun}}, \binits{A.}},
\bauthor{\bsnm{{Lee}}, \binits{Y.}}:
\batitle{Performance analysis of sensor systems for space situational awareness}.
\bjtitle{Journal of Astronomy and Space Sciences}
\bvolume{34},
\bfpage{303}--\blpage{314}
(\byear{2017})
\doiurl{10.5140/JASS.2017.34.4.303}
\end{barticle}
\endbibitem

\bibitem[\protect\citeauthoryear{Agency}{2024}]{darpa_sst}
\begin{botherref}
\oauthor{\bsnm{Agency}, \binits{D.A.R.P.}}:
Space Surveillance Telescope (SST).
Accessed: 2024-06-25
(2024).
\url{https://www.darpa.mil/program/space-surveillance-telescope}
\end{botherref}
\endbibitem

\bibitem[\protect\citeauthoryear{Harrison et~al.}{2021}]{Harrison2021}
\begin{botherref}
\oauthor{\bsnm{Harrison}, \binits{T.}},
\oauthor{\bsnm{Johnson}, \binits{K.}},
\oauthor{\bsnm{Roberts}, \binits{T.G.}}:
Defense Against the Dark Arts in Space: Protecting Space Systems from Counterspace Weapons.
\url{https://csis-website-prod.s3.amazonaws.com/s3fs-public/publication/210225_Harrison_Defense_Space.pdf?N2KWelzCz3hE3AaUUptSGMprDtBlBSQG}
\end{botherref}
\endbibitem

\bibitem[\protect\citeauthoryear{Vinti}{1966}]{Vinti_1966}
\begin{barticle}
\bauthor{\bsnm{Vinti}, \binits{J.P.}}:
\batitle{Invariant properties of the spheroidal potential of an oblate planet}.
\bjtitle{Journal of Research of the National Bureau of Standards}
\bvolume{70B},
\bfpage{1}--\blpage{16}
(\byear{1966})
\end{barticle}
\endbibitem

\bibitem[\protect\citeauthoryear{Biria and Russell}{2018}]{Biria_2018}
\begin{barticle}
\bauthor{\bsnm{Biria}, \binits{A.D.}},
\bauthor{\bsnm{Russell}, \binits{R.P.}}:
\batitle{Equinoctial elements for {V}inti theory: Generalizations to an oblate spheroidal geometry}.
\bjtitle{Acta Astronautica}
\bvolume{153},
\bfpage{274}--\blpage{288}
(\byear{2018})
\doiurl{10.1016/j.actaastro.2017.11.013}
\end{barticle}
\endbibitem

\bibitem[\protect\citeauthoryear{Jones and Weisman}{2019}]{Jones2019}
\begin{barticle}
\bauthor{\bsnm{Jones}, \binits{B.A.}},
\bauthor{\bsnm{Weisman}, \binits{R.}}:
\batitle{Multi-fidelity orbit uncertainty propagation}.
\bjtitle{Acta Astronautica}
\bvolume{155},
\bfpage{406}--\blpage{417}
(\byear{2019})
\doiurl{10.1016/j.actaastro.2018.10.023}
\end{barticle}
\endbibitem

\bibitem[\protect\citeauthoryear{Vo and Vo}{2012}]{Vo2012}
\begin{bchapter}
\bauthor{\bsnm{Vo}, \binits{B.T.}},
\bauthor{\bsnm{Vo}, \binits{B.N.}}:
\bctitle{The para-normal {B}ayes multi-target filter and the spooky effect}.
In: \bbtitle{2012 15th International Conference on Information Fusion},
pp. \bfpage{173}--\blpage{180}
(\byear{2012})
\end{bchapter}
\endbibitem

\end{thebibliography}

\end{document}